\documentclass[letterpaper,twocolumn,10pt]{article}

\usepackage{times} 
\usepackage{authblk} 
\usepackage{usenix-2020-09}
\usepackage{pifont}
\usepackage{balance}
\usepackage{enumitem}
\usepackage{graphicx}
\usepackage{subfigure}
\usepackage{multirow}
\usepackage{bbm}

\usepackage{tablefootnote}
\usepackage{threeparttable}
\usepackage{xspace}
\usepackage[ruled,vlined,linesnumbered,ruled]{algorithm2e}
\usepackage{amsmath}
\usepackage{amsfonts}
\usepackage{amssymb}
\usepackage{bm}
\usepackage{booktabs}
\usepackage{makecell}
\usepackage{setspace}

\def\ie{\textit{i.e.}\xspace}

\def\etc{\textit{etc.}\xspace}

\def\eg{\textit{e.g.}\xspace}

\def\method{FedSpine\xspace}

\usepackage{tikz}
\usepackage{amsmath}

\usepackage{filecontents}

\makeatletter
\renewcommand{\maketag@@@}[1]{\hbox{\m@th #1}}
\makeatother

\begin{document}

\date{}

\title{\Huge \bf Efficient Deployment of Large Language Models on Resource-constrained Devices}

\author{
    \makebox[0.9\textwidth][c]{Zhiwei Yao$^{\dag,\ddag}$ \ \
    Yang Xu$^{\dag,\ddag}$ \ \
    Hongli Xu$^{\dag,\ddag}$ \ \
    Yunming Liao$^{\dag,\ddag}$ \ \
    Zuan Xie$^{\dag,\ddag}$} \\
    \makebox[0.9\textwidth][c]{\textsuperscript{\dag}School of Computer Science and Technology, University of Science and Technology of China} \\ 
    \makebox[0.9\textwidth][c]{\textsuperscript{\ddag}Suzhou Institute for Advanced Research, University of Science and Technology of China} \\ 
    \makebox[0.9\textwidth][c]{\{zhiweiyao, ymliao98, xz1314\}@mail.ustc.edu.cn, \{xuyangcs, xuhongli\}@ustc.edu.cn}
}

\maketitle

\begin{abstract}

Deploying Large Language Models (LLMs) on resource-constrained  (or weak) devices presents significant challenges due to limited resources and heterogeneous data distribution. 
To address the data concern, it is necessary to fine-tune LLMs using on-device private data for various downstream tasks.
While Federated Learning (FL) offers a promising privacy-preserving solution, 
existing fine-tuning methods retain the original LLM size, 
leaving issues of high inference latency and excessive memory demands unresolved.
Hence, we design \method, an FL framework that combines Parameter-Efficient Fine-Tuning (PEFT) with structured pruning for efficient deployment of LLMs on resource-constrained devices.
Specifically, \method introduces an iterative process to 
prune and tune the parameters of LLMs.
To mitigate the impact of device heterogeneity,  
an online Multi-Armed Bandit (MAB) algorithm is employed to adaptively determine different pruning ratios and LoRA ranks for heterogeneous devices without any prior knowledge of their computing and communication capabilities.
As a result, \method maintains higher inference accuracy while improving fine-tuning efficiency.
Experimental results conducted on a physical platform with 80 devices demonstrate that \method can speed up fine-tuning by 1.4$\times$-6.9$\times$ and improve final accuracy by 0.4\%-4.5\% under the same sparsity level compared to other baselines.
\end{abstract}

\section{Introduction}
The recent development of Large Language Models (LLMs), such as OpenGPT and LLaMA, marks a significant milestone in advancing Artificial Intelligence (AI) \cite{cao2023comprehensive}. 
As computational capabilities on end devices continue to improve, leading technology companies, including Apple and OpenAI, have introduced LLM-powered and on-device AI applications, such as Apple Intelligence \cite{gunter2024apple} and ChatGPT mobile apps.
These applications are usually driven by a small model (\eg, with around 3 billion parameters) for on-device inference, and a much larger server-side model (\eg, the 175 billion parameter GTP-3 model) for more complicated tasks.
The small on-device model is usually pre-trained or distilled from a larger model on the server-side with in-cloud centralized data.
However, the heterogeneous and context-specific data generated on end devices often exhibits distributional shifts from the data on which LLMs are pre-trained, leading to reduced inference accuracy.
Besides, the limited resources (\eg, computational and memory resources) of the ``weak'' devices inevitably restrict the sizes of on-device models.
\textit{Therefore, the data and resource properties hinder the deployment of LLMs on weak devices for effective and efficient inference.}

To address the data concern, fine-tuning LLMs with on-device data helps the deployed models to better align with the unique requirements of downstream tasks, thereby enhancing task-specific performance and adaptability.
However, growing privacy concerns and stringent regulations, such as GDPR \cite{voigt2017eu}, restrict the collection and sharing of personal raw data, making it challenging to gather distributed data from end devices to centralized servers for LLM fine-tuning. 
To break this barrier, Federated Learning (FL) has emerged as a privacy-preserving distributed machine learning technique that orchestrates end devices to collaboratively train/fine-tune models without exposing their raw data \cite{mcmahan2017communication, wang2018edge}.
When it comes to LLMs, directly conducting full parameter fine-tuning with FL is challenging  due to the limited resources on the devices. 
For instance, fine-tuning a LLaMA model with 13 billion parameters requires approximately 100GB of memory, while most end devices have only 4-12GB of memory available \cite{xu2024fwdllm}.
To cope with the resource issue, integrating Parameter-Efficient Fine-Tuning (PEFT) techniques within FL presents a compelling solution. 
PEFT methods, such as Adapter \cite{houlsby2019parameter} and LoRA \cite{hu2021lora}, aim to freeze the pre-trained LLM parameters and update only a few additional parameters (typically less than 1\% of total LLM parameters) \cite{zhang2023fedpetuning}, facilitating the implementation of fine-tuning LLMs on weak devices.

While PEFT methods effectively address resource constraints during the fine-tuning process, the resulting models often retain or even increase their original size due to the introduction of additional parameters. 
This will lead to significant inference latency or even unresponsiveness on resource-constrained devices, as the computational and memory demands during inference remain high. 
To alleviate these issues, model pruning has emerged as a promising technique to reduce the number of model parameters and computational footprint.
Pruning techniques can be classified into unstructured pruning and structured pruning.  
Unstructured pruning produces highly sparse models \cite{han2015deep, sanh2020movement} but hardly provides inference speedup without specialized hardware support.
In contrast, structured pruning \cite{ma2023llm, kwon2022fast, xia2022structured, zhao2024apt}, which removes groups of structured parameters such as attention heads and feed-forward neurons (FFN),  significantly decreases model size, improves inference speed and lowers memory usage.  In this work, we focus on adopting a structured pruning approach. 

Intuitively, integrating PEFT and structured pruning offers promise for enhancing both accuracy and efficiency of on-device LLM inference. 
However, recent studies \cite{zhao2023cpet, ma2023llm} indicate that directly combining PEFT with structured pruning will lead to significant performance degradation, as the two techniques have not been jointly optimized. 
For example, CPET \cite{zhao2023cpet} applies structured pruning to models fine-tuned with LoRA, while LLMPruner \cite{ma2023llm} adopts a sequential approach: first pruning the model, and then using LoRA fine-tuning to restore performance. 
However, this separation of pruning and fine-tuning results in suboptimal performance \cite{sanh2020movement, molchanov2019importance}, as verified in \S \ref{sec: combination of pruning and fine-tuning}.
To address these limitations, APT \cite{zhao2024apt} introduces an adaptive framework that iteratively prunes and fine-tunes the parameters of LLMs, achieving significant improvements in both fine-tuning and inference performance.
However, these studies including APT are specialized and implemented for centralized scenarios in the cloud, which raises significant privacy concerns due to the need for data collection from end devices. 
Moreover, they always incur substantial computational and memory overheads, which cannot be directly applied to end devices considering their limited resources.

To this end, we propose \method, a novel and the first (to our best knowledge) FL framework that combines structured pruning and PEFT to enhance inference efficiency of on-device LLMs. 
Specifically \method introduces an iterative process to prune and fine-tune model parameters. In each round, devices download the updated LoRA weights from the server and perform local pruning on the frozen model. Afterward, the pruned model is fine-tuned on local data and the updated weights are sent back to the server for aggregation.
However, apart from resource limitation, one critical issue complicates the design of \method, \ie, device heterogeneity. 
Specifically, end devices are typically equipped with different computational chips and located in diverse regions, causing their computational and communication capabilities to vary significantly, even by more than tenfold times \cite{chen2022decentralized, lai2021oort}. 
If identical pruning and fine-tuning settings, such as uniform pruning ratios and LoRA ranks, are assigned across devices, those with weak capabilities may become system stragglers \cite{fang2024automated}, resulting in considerable competition time lags and decreased fine-tuning efficiency.
Thus, it is worthwhile to design effective strategies to mitigate the impact of device heterogeneity.
To cope with these issues, \method adaptively assigns appropriate pruning ratios and LoRA ranks for devices in each round to mitigate the straggler effect caused by device heterogeneity.
Regarding the observations in \S \ref{sec: motivation}, the potential accuracy degradation incurred by structured pruning of different ratios could be compensated by adaptive assignment of LoRA ranks.   
Higher LoRA ranks help to achieve accuracy gains but come with the cost of increased resource overhead (\eg, memory usage and communication time). 
Conversely, lower LoRA ranks reduce resource consumption but may compromise accuracy recovery. 
Therefore, \method should jointly determine proper pruning ratios and LoRA ranks across devices through iterative optimization so as to maximize the fine-tuning efficiency and minimize accuracy loss. 

In a nutshell, our work makes the following contributions: 
\begin{itemize}
    \item We propose \method, an innovative federated fine-tuning framework that enhances both inference accuracy and efficiency of on-device LLMs.  
    This framework integrates adaptive pruning on the frozen model and LoRA ranks on the LoRA modules to effectively address the challenges posed by device heterogeneity. 
    
    \item \method employs a Multi-Armed Bandit (MAB) based online learning algorithm to adaptively learn the quantitative relationship between resource consumption and model performance. 
    The algorithm determines proper pruning ratios and LoRA ranks for heterogeneous devices without any prior knowledge of their capabilities.

    \item We conduct comprehensive experiments to evaluate the performance of \method on a physical platform consisting of 80 NVIDIA Jetson devices. 
    The results show that \method speeds up fine-tuning by 1.4$\times$-6.9$\times$ and improves accuracy by 0.4\%-4.5\% compared to baselines. 
\end{itemize}

\vspace{-4mm}


\section{Preliminaries and Motivations}\label{sec:prelim}

\vspace{-1mm}
\subsection{Integrating LoRA into FL: Benefits and Potential Challenges}\label{subsec: model prune}

\begin{figure}[!t]
\centering
    \includegraphics[width=0.46\textwidth]{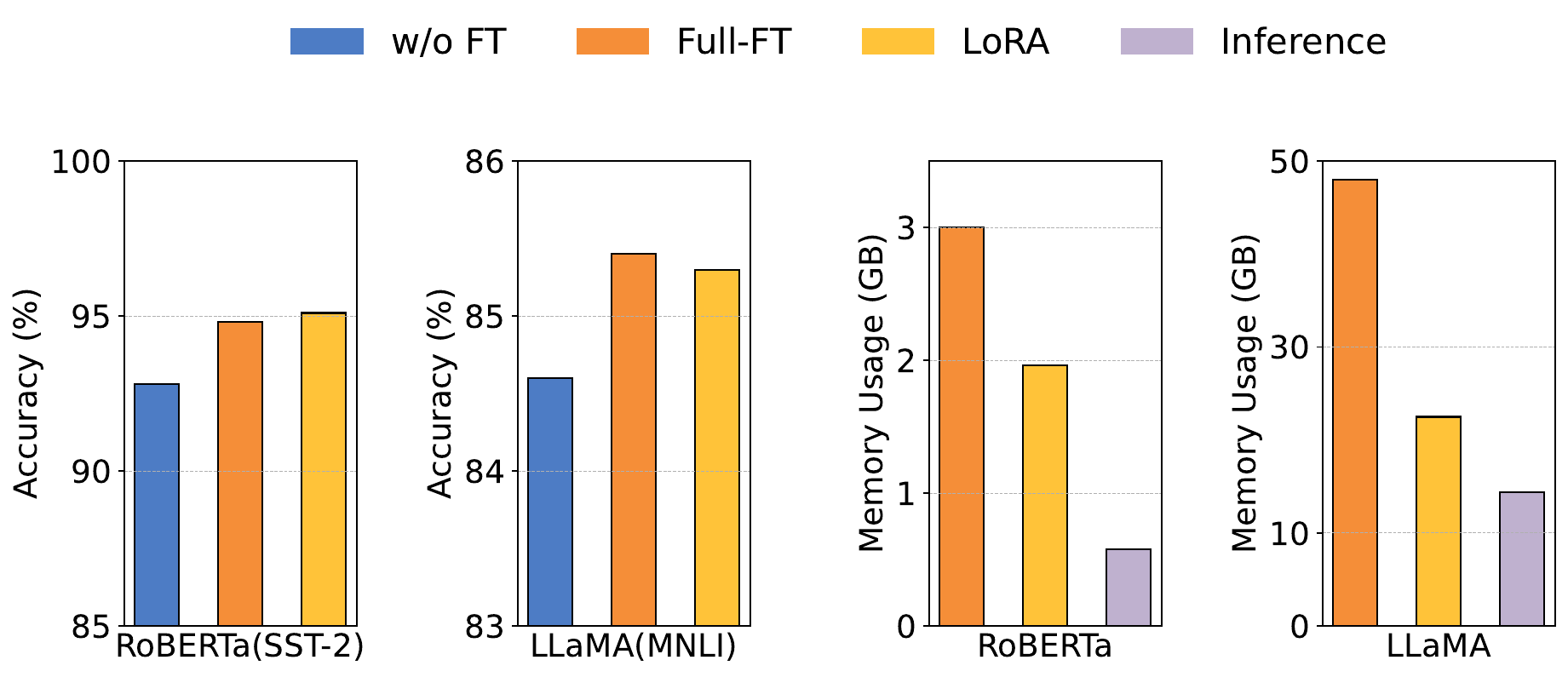}
    \vspace{-2mm}
\caption{Peak memory footprint and test accuracy of different fine-tuning methods and inference.} 
\label{fig: accuracy and memory}
\vspace{-4mm}
\end{figure}

LLMs have demonstrated remarkable capabilities in language understanding and generation.
Before deployment, two critical steps are involved: 1) pre-training LLMs from scratch with extensive computational resources and tremendous text corpora; and 2) fine-tuning pre-trained LLMs to fit various downstream tasks. 
With the utilization of on-device data, fine-tuning allows LLMs to align more closely with specific task requirements, thereby enhancing task-specific performance. 
As depicted in Figure \ref{fig: accuracy and memory}, full parameter fine-tuning (Full-FT) for RoBERTa on SST-2 can improve accuracy of 2\%-3\% compared to direct deploying LLMs on end devices without fine-tuning (w/o FT). 

However, performing Full-FT on resource-constrained devices within standard FL frameworks is challenging due to the extensive parameters of LLMs, which leads to significant computational and communication overhead in both model tuning and aggregation. 
To address this, PEFT methods have emerged as potential solutions that could obtain comparable performance to Full-FT with reduced resource costs.
Among these methods, LoRA \cite{hu2021lora} stands out as the most popular fine-tuning technique for LLMs.
Specifically, LoRA injects trainable low-rank decomposition matrices into each layer of an LLM, while freezing all pre-trained parameters of the LLM.
By selectively updating and transmitting the trainable low-rank matrices, LoRA seeks to enhance fine-tuning efficiency. 

\begin{table}[!t]
\caption{Inference performance of LLaMA-7B with different pruning ratios on SST-2.}
\label{table: inference incurred by pruning}
\centering
\scalebox{0.96}{
    \begin{tabular}{c|ccccc}
    \hline
    Pruning ratio & 0 & 0.1 & 0.2 & 0.3 & 0.4  \\ 
    \hline
    Inf. Time (ms)  & 62.5 & 56.0 & 49.8 & 43.5 & 37.3  \\ 
    Inf. Memory (GB) & 17.6 & 17.0 & 16.5 & 15.9 & 15.4 \\
    \hline
    \end{tabular}
}
\vspace{-0.4cm}
\end{table}

To evaluate the effectiveness of LoRA for federated fine-tuning of LLMs, we compare its performance with that of Full-FT on RoBERTa and LLaMA-7B.
As shown in Figure \ref{fig: accuracy and memory}, LoRA-based fine-tuning brings 34.7\%-46.2\% memory usage savings across different models.
Figure \ref{fig: computation and communication overhead} shows the fine-tuning time on a batch of data and the amount of exchanged parameters for both methods.
We observe that LoRA-based fine-tuning achieves more than 2.1$\times$ speedup in training, and reduces the amount of exchanged parameters by 50$\times$ than Full-FT.
These results underscore LoRA's potential to mitigate computational and communication bottlenecks, making it a viable candidate for incorporation into FL to save resources. 

Nevertheless, integrating FL with LoRA for LLMs still poses potential challenges.
Fine-tuned models often retain or even increase in model size due to the additional parameters introduced, leaving issues of high inference latency and excessive memory demands unresolved.  
As shown in Figure \ref{fig: accuracy and memory},  peak memory usage during inference remains substantial, even with a small batch size of 8, often exceeding the capacity of resource-constrained devices.
For example, LLaMA-7B inference can surpass the memory limits of most end devices, which typically offer only 4-12GB of memory \cite{xu2024fwdllm}.
Moreover, all existing federated fine-tuning methods fail to accelerate inference, underscoring the need to reduce model size for effective deployment on resource-constrained devices.

  
    


\vspace{-1mm}
\subsection{Combination of Pruning and Fine-tuning} \label{sec: combination of pruning and fine-tuning}
Structured pruning for LLMs \cite{zhao2023cpet, ma2023llm, zhao2024apt, zhang2023loraprune} focuses on removing consistent blocks in transformer layers (\eg, attention heads, FFN neurons, and hidden dimensions).
As a result, the resulting model contains fewer parameters, improving inference efficiency and reducing memory usage.
Table \ref{table: inference incurred by pruning} illustrates the inference benefits of structured pruning by showing the inference latency and memory usage of LLaMA-7B on SST-2 with a batch size of 8 by varying pruning ratios.
As the pruning ratio increases, the model size becomes smaller and can achieve more inference speedup and memory savings.

Recent works have explored combining fine-tuning and structured pruning to optimize LLM deployment on devices.
Methods such as LLMPruner \cite{ma2023llm} and CPET \cite{zhao2023cpet} adopt different sequential strategies to combine these techniques.
LLMPruner begins by pruning the model and subsequently applying LoRA-based fine-tuning to recover its performance. 
In contrast, CPET  reverses the sequence, performing LoRA-based fine-tuning before applying structured pruning.
However, both methods rely on a strict separation of pruning and fine-tuning stages, often leading to suboptimal performance \cite{sanh2020movement, molchanov2019importance}. 
To conquer this issue, APT \cite{zhao2024apt} introduces a novel framework that alternates between pruning and fine-tuning iteratively, resulting in notable improvements in both fine-tuning and inference performance.
Nevertheless, APT is designed specifically for centralized cloud settings,  making it unsuitable for deployment on resource-constrained devices.
%
\begin{figure*}[t]
        
	\centering
	\subfigure[ ]{
		\centering
		\includegraphics[width=0.253\linewidth]{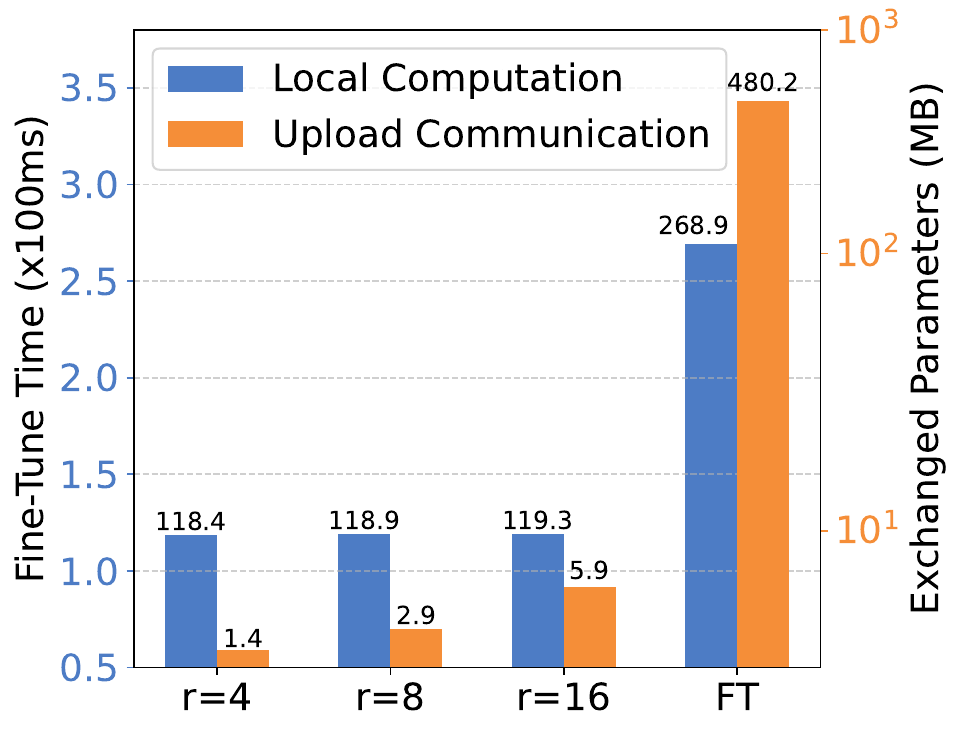}\label{fig: computation and communication overhead}
	}
	\subfigure[ ]{
		\centering
		\includegraphics[width=0.23\linewidth]{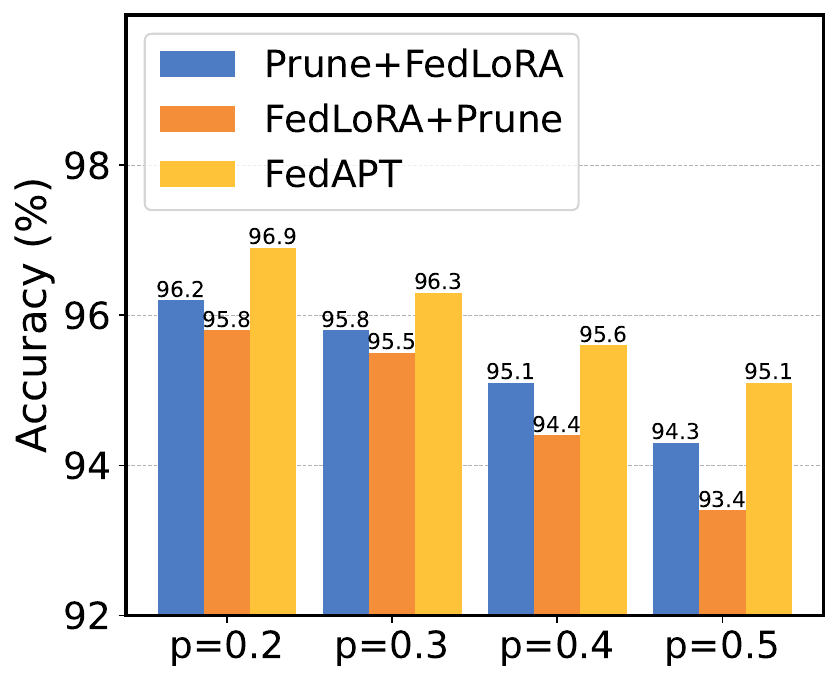}\label{fig: iterative optimization}
	}
        \subfigure[ ]{
		\centering
		\includegraphics[width=0.226\linewidth]{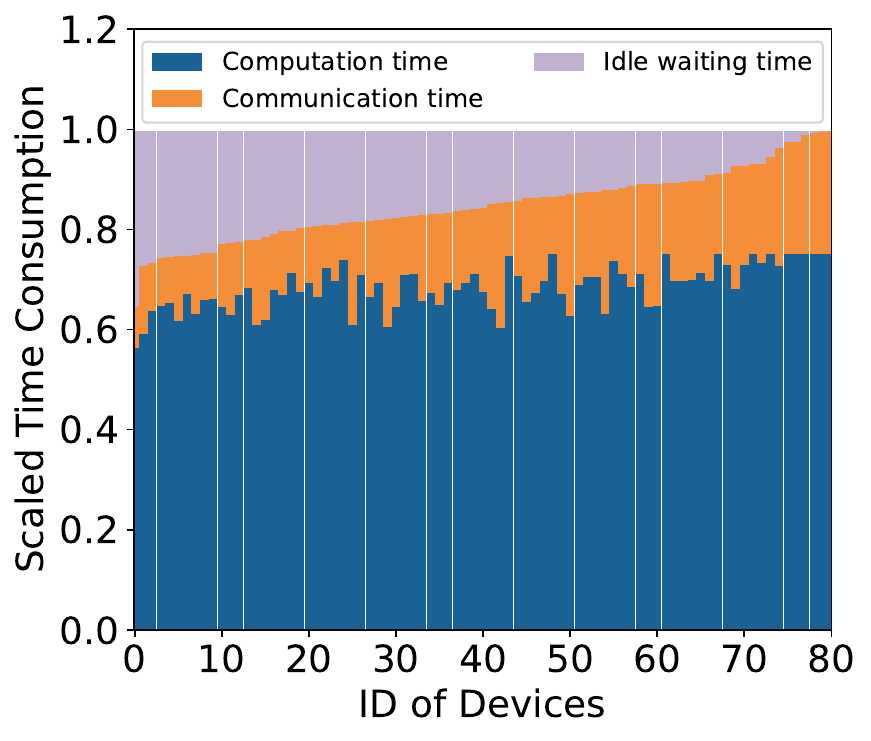}\label{fig: stragger_time}
        }
        \subfigure[ ]{
		\centering
		\includegraphics[width=0.226\linewidth]{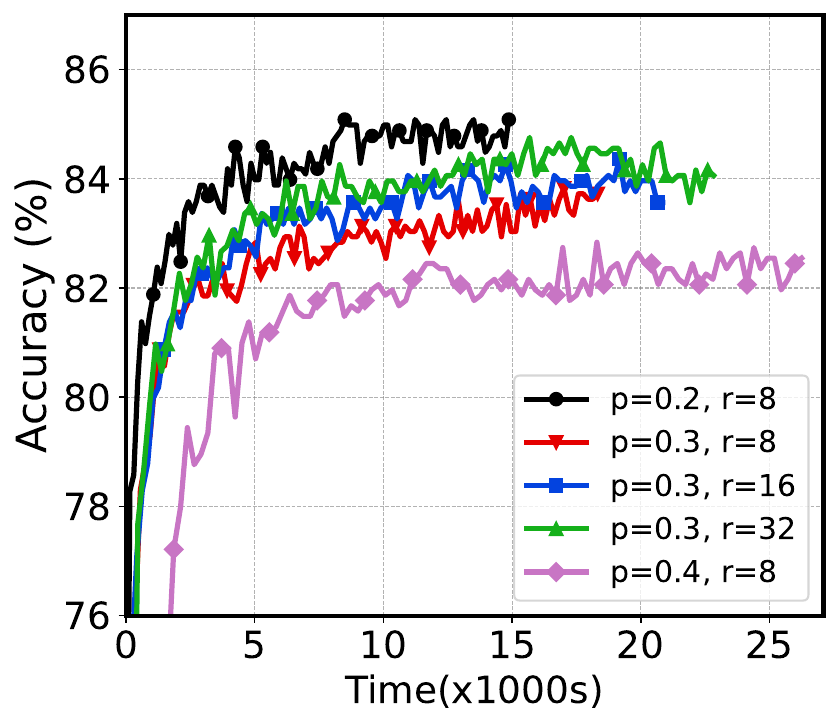}\label{fig: adaptive pruning and tuning}
	}
        \vspace{-1mm}
	\caption{The results of preliminary experiments ($p$ and $r$ separately denote pruning ratio and LoRA rank). (a) The fine-tuning time and exchanged parameters of LoRA and Full-FT for RoBERTa on SST-2; (b) Test accuracy of three methods after reaching the target pruning ratio on SST-2; (c) Ranked completion time consumed by heterogeneous devices with identical pruning ratios and LoRA ranks in one round; (d) The fine-tuning process of FedAPT with different pruning ratios and LoRA ranks.}\label{fig:preliminary results}
        \vspace{-3mm}
\end{figure*}

To demonstrate the effectiveness of iterative optimization of pruning and fine-tuning, we compare three approaches: 1) Prune+FedLoRA: pruning the model first, followed by federated LoRA-based fine-tuning, 2) FedLoRA+Prune: federated LoRA-based fine-tuning followed by pruning,
and 3) an iterative approach (\ie, APT) adapted to federated settings, referred to as FedAPT.
In each round of FedAPT, each device downloads the updated LoRA weights from the server and performs several fine-tuning iterations (\eg, 20). This is followed by a local model pruning process, after which the updated LoRA weights are sent back to the server for aggregation.
For these experiments, we adopt the same pruning strategy as in APT, \ie, gradient-guided pruning, to efficiently estimate the importance of pre-trained weights.
All participating devices are assigned the same pruning ratio.
The results, as shown in Figure \ref{fig: iterative optimization}, demonstrate that FedAPT achieves superior performance compared to other methods regardless of the pruning ratio. 
For example, when the pruning ratio is 0.3 for all devices, FedAPT improves the final accuracy by 0.5\%-1.7\%  compared with other methods. 
This highlights the importance of iterative optimization in leveraging the benefits of both pruning and fine-tuning in FL.
Therefore, we are inspired to design an FL
framework to iteratively perform structured pruning while simultaneously conducting efficient fine-tuning.   
Although FedAPT demonstrates superior performance compared to other methods, 
it may fail to cope with device heterogeneity, which has been validated in \S \ref{sec: device heterogeneity}.

\vspace{-1.8mm}
\subsection{Impact of Device Heterogeneity} \label{sec: device heterogeneity}
Due to device heterogeneity, computational and communication time will vary significantly across devices in each round. 
When identical pruning ratios are adopted, devices with limited memory and computational capabilities become bottlenecks during model fine-tuning. 
This forces strong devices to wait for weak ones (also called stragglers) for global LoRA aggregation, significantly reducing fine-tuning efficiency. 
Moreover, most PEFT-based federated fine-tuning schemes \cite{zhang2023fedpetuning, yang2024dual, yi2023fedlora} assign identical and fixed LoRA ranks to all devices, overlooking the impact of device heterogeneity.

To evaluate the negative impacts of device heterogeneity, we conduct experiments with  FedAPT for fine-tuning RoBERTa on MNLI \cite{wang2018glue} using 80 NVIDIA Jetson devices.
Specifically, we select three types of devices (\ie, AGX, TX2, and NX devices) and configure them to operate in different modes and network bandwidths to simulate heterogeneous computational and communication capabilities (see \S \ref{sec:evaluation} for more details).
We record the ranked completion time of all devices 
in a round, with identical LoRA ranks (\ie, 8) and pruning ratios (\ie, 0.2) assigned.
As shown in Figure \ref{fig: stragger_time}, there is a significant disparity in computational and communication time across devices.
The strongest device requires only 70\% of the time taken by the weakest device to complete one round, resulting in approximately 30\% of the strongest device’s time being idle and inefficiently utilized.

\vspace{-1mm}
\subsection{Motivations for Adaptive Pruning and Fine-tuning} \label{sec: motivation}
\vspace{-0.5mm}
To further explore the effect of different pruning ratios and LoRA ranks on fine-tuning, we conduct a set of pre-experiments using FedAPT for fine-tuning the RoBERTa model on MNLI over 100 rounds.
The fine-tuning processes are presented in Figure \ref{fig: adaptive pruning and tuning} 
and we can observe the key findings as follows. 
1) Larger pruning ratios can lead to performance degradation and slower model convergence.
For example, increasing the pruning ratio from 0.2 to 0.3 results in a 1.4\% drop in accuracy and requires additional time for convergence.
2) Larger LoRA ranks help recover model performance but come with more per-round time consumption. 
Under the same pruning ratio of $p$ = 0.3, models with a larger LoRA rank ($r$ = 32) achieve 0.9\% higher accuracy but require approximately 23.3\% more total time across all rounds compared to models with a lower LoRA rank ($r$ = 8).

The above results highlight the importance of carefully determining pruning ratios and LoRA ranks, underscoring  the need for joint optimization. 
To deal with device heterogeneity and minimize the straggler effects during fine-tuning, it is essential to dynamically assign different pruning ratios and LoRA ranks based on the memory, computational and communication capabilities of heterogeneous devices.
However, determining the optimal pruning ratios and LoRA ranks remains challenging, as it demands balancing the inference accuracy and fine-tuning efficiency, which is the primary objective of our method.

\vspace{-2mm}

\section{System Overview }\label{sec:algorithm}
\begin{figure*}[!t]
\centering
    \includegraphics[width=0.90\textwidth]{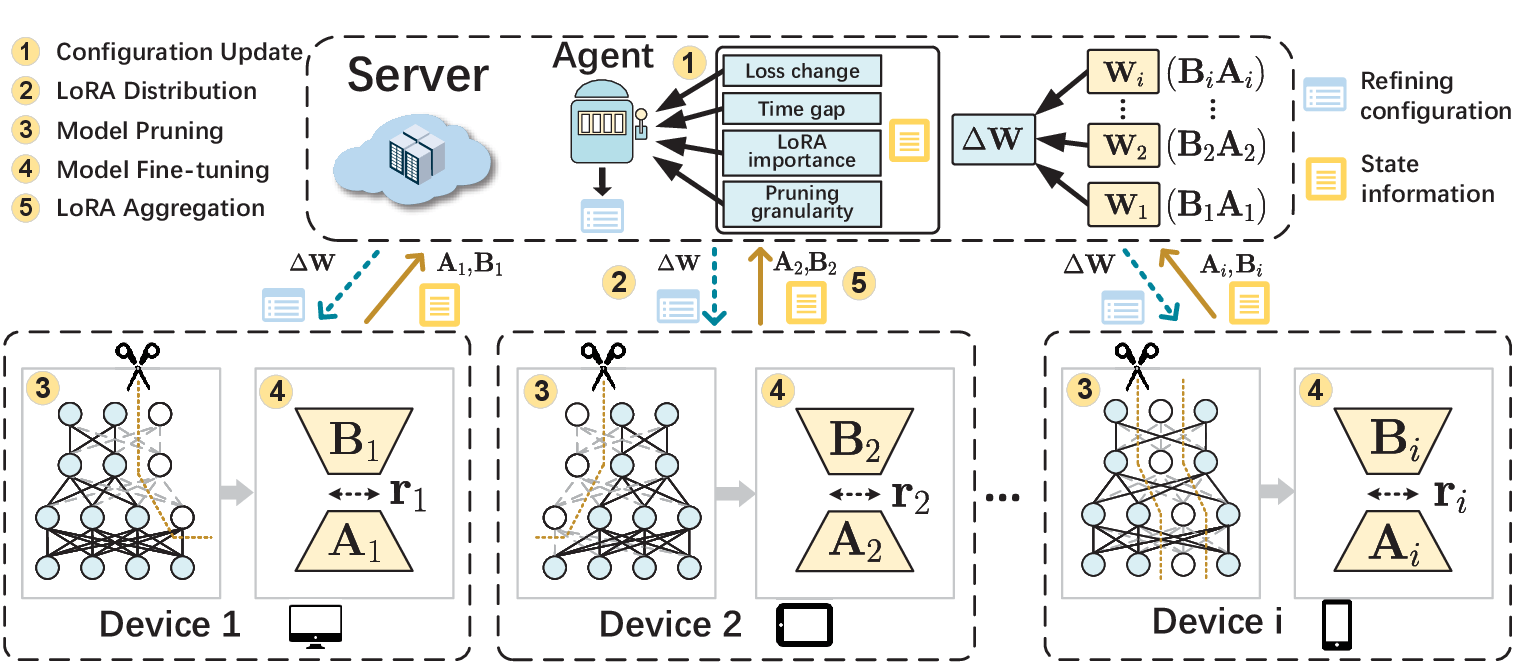}
    \vspace{-3mm}
\caption{Overview of \method's workflow.} 
\label{fig: System}
\vspace{-3mm}
\end{figure*}

\vspace{-1mm}
In this section, 
we introduce the overall workflow of \method (as illustrated in Figure \ref{fig: System}). 
The working procedure of \method typically involves a certain number of rounds.
Each round contains five main stages to ``refine'' the LLMs as follows, which are performed iteratively until the on-device LLMs are progressively pruned to the target pruning ratio.

\textcircled{1} \textbf{Configuration Update.} 
At the beginning of round $t$ ($t\geq1$), the server utilizes the state information  collected in the previous round (\eg, local loss changes) to adaptively determine  the refining configurations (\ie, LoRA ranks and pruning ratios) for each device in the current round(\S \ref{sec: Configuration Update}).

\textcircled{2} \textbf{LoRA Distribution.} 
The server distributes the LoRA weights updated in the previous round, and the refining configurations to all devices.





\textcircled{3} \textbf{Model Pruning.} 
Devices efficiently estimate the importance of the weights in their original LLMs by employing the gradients of LoRA obtained in the previous round.
According to the received refining configurations, unimportant weights are progressively pruned until the designated pruning ratio is reached (\S \ref{sec: model pruning}).

\textcircled{4} \textbf{Model Fine-tuning.}
Each device shrinks the received LoRA weights based on its assigned LoRA rank and combines them with its local LLM to assemble a full model.
The assembled model is then fine-tuned on the local dataset for several iterations while  recording key state information. 
Once pruning and fine-tuning are complete, the updated LoRA weights and recorded state information are transmitted back to the server for aggregation and configuration updates (\S \ref{sec: model fine-tuning}). 

%
         

\textcircled{5} \textbf{LoRA Aggregation.} Upon receiving the LoRA weights from all devices, the server performs adaptive global aggregation. 
In \method, we adopt a heterogeneity-aware weighted aggregation strategy to effectively handle the heterogeneous LoRA ranks across devices (\S \ref{sec: LoRA aggregation}). 

\vspace{-2mm}

\section{\method Design}\label{sec:algorithm}
\vspace{-2mm}
Herein, we will elaborate on the technical details of four critical stages, \ie, Configuration Update, Model Pruning, Model Fine-tuning, and LoRA Aggregation.
\vspace{-1mm}
\subsection{Algorithm for Configuration Update}  \label{sec: Configuration Update}
\vspace{-1mm}
In this section, we first formalize the problem in \method and then propose an online learning algorithm to adaptively determine the pruning ratios and LoRA ranks 
for devices.
\subsubsection{Problem Definition}
In \method, there are $N$ devices, each device $i \in [1, N]$ keeping its local data $\mathcal{D}_i$.
For a frozen weight matrix $\mathbf{W}_i^0 \in \mathbb{R}^{d \times k}$ of LLMs on device $i$,  LoRA constrains its update $\Delta \mathbf{W} \in \mathbb{R}^{d \times k}$ by representing it with a low-rank decomposition $\mathbf{W}_i^0 + \Delta \mathbf{W} = \mathbf{W}_i^0 + \mathbf{B} \mathbf{A}$, where 
$\mathbf{B} \in \mathbb{R}^{d \times r}$ and $\mathbf{A} \in \mathbb{R}^{r \times k}$ are two globally trainable parameters.
Here, $d$ and $k$ denote the input and output dimensions of the frozen weight matrix, and the rank $r$ satisfies $r \ll \min (d, k)$. 
The completion time $T_i^t$ of device $i$ in round $t$ includes local computational time $T_{i,\text{comp}}^t$ and communication time $T_{i,\text{comm}}^t$, \ie, $T_i^t = T_{i,\text{comp}}^t + T_{i,\text{comm}}^t$, both influenced by the pruning ratios and LoRA ranks. The total latency for each round is determined by the "weakest" device, expressed as $\max_i\{T_i^t\}$, resulting in idle time for stronger devices. The average waiting time across all devices is:
\begin{equation} \label{Eq: waiting time}
\varGamma^t=\frac{1}{N} \sum_{i=1}^N(\max \{T_i^t\}-T_i^t)
\vspace{-2mm}
\end{equation} 

To conquer these inefficiencies, the optimization problem in FedSpine aims to minimize the global loss $F$ while ensuring that device parameters achieve a target pruning ratio $p_i^T$ after $T$ fine-tuning rounds and maintaining efficiency constraints:
\vspace{-2mm}

\begin{algorithm}[!t]
	\caption{Smooth Upper Confidence Bound for device $i$} \label{alg: S-UCB}
		\For{Each round $t= \{1, 2, \dots,T \}  $}
		{
			\For {Each partition region $S_j^t \in \mathbf{S^t}$}
			{   
				Calculate upper confidence bound $U_t(S_j^t)=\bar{R}_t(\lambda, S_j^t)+c_t(\lambda, S_j^t)$\;
				Choose the partition region $ S_j^t=  U_t(S_j^t) $\; 
                Select the pruning ratio $p_i^t$ and $r_i^t$ from the $S_j^t$\; 
				\If{The diameter of $S_j^t$ is larger than $\delta$}{
					Split $S_j^t$ with $p_i^t$ and $r_i^t$ to form $S^{t+1}$\;}
                \Else{$S^{t+1}$ = $S^t$\;}
                }
                
				Record the completion time $T_i^t$ of device $i$\;
				Calculate the reward $R (p_i^t, r_i^t)$;
	}
\end{algorithm}

$$\underset{\Delta \mathbf{W}}{\operatorname{min}} \frac{1}{N} \sum_{i=1}^N \frac{1}{\mid \mathcal{D}_i\mid} \sum_{d_i  \in \mathcal{D}_i} F_i(\mathbf{W}_i; \Delta \mathbf{W}; d_i  ) $$
\vspace{-1.5mm}
\begin{equation}\label{eq: problem} 
    s.t.
    \begin{cases}
    1 - \frac{\texttt{ \normalsize size}(\mathbf{W}_i^t)}{\texttt{ \normalsize size}  (\mathbf{W}_i^0)} \geq p_i^t  \hspace{0.7em} \hfill  \triangleright \text { Preference Constraint } \\
    
    \varGamma^t \leq \epsilon \hfill \triangleright \text { Efficiency Constraint }\\
     \forall t \in [0, T], \forall i \in [1, N]  \\ 
     
    \end{cases}
\end{equation}
where $F_i$ is local loss function, $d_i $ is a batch of data samples in $\mathcal{D}_i$ and $\epsilon$ is a positive constant  close to zero. 
According to Eq. \eqref{eq: problem}, a higher pruning ratio reduces computational and memory demands during inference but sacrifices inference accuracy.
Conversely, fine-tuning with a larger LoRA rank requires more time and memory usage but accelerates model convergence with better performance.  
\method balances inference accuracy and fine-tuning efficiency by assigning different pruning ratios and LoRA ranks.

\vspace{-1mm}
\subsubsection{Multi-Armed Bandit Based Algorithm} \label{subsec: MAB}
\vspace{-1mm}
Due to the complex and varying nature of federated environments (\eg, heterogeneous device capabilities), it is infeasible to predefine the optimal values of the pruning ratio and LoRA rank for all devices in each round.
Therefore, we propose a Multi-Armed Bandit (MAB) based online learning algorithm to dynamically determine the model refining configurations, \ie, pruning ratios and LoRA ranks, for heterogeneous devices in real-time without any prior knowledge of their capabilities.

Generally, the optimization problem in Eq. \eqref{eq: problem} can be modeled as an MAB problem, where the server acts as the player, and the decisions of pruning ratio and LoRA rank represents the arms. 
In each round $t$, the server decides which arm of the bandit is pulled, and a reward in Eq. \eqref{eq: reward} will be collected after the decision.
The Upper Confidence Bound (UCB) policy is commonly used to address Multi-Armed Bandit (MAB) problems \cite{gao2020combinatorial}. 
Traditionally, UCB was developed for discrete decision spaces. 
However, in \method, the pruning ratios form a continuous range, varying between [0,1].
To overcome this, we extend the UCB policy to handle continuous arms and introduce the Smooth Upper Confidence Bound (S-UCB), as described in Algorithm \ref{alg: S-UCB}.

Specifically, S-UCB creates agents for different devices and a decision tree is used to adaptively learn arm space partitions. 
Considering the progressive pruning process in \method, the pruning ratio $p^t$ in round $t$ must exceed the ratio $p^{t-1}$ from the previous round.
In round $t$, each agent maintains a sequence of finite partitions of the arm space $\mathbf{S}^t=\{S_1^t, S_2^t, \ldots, S_{l_t}^t\}$ with $\cup_{j=1}^{l_t} S_j^t= [p^{t-1},p^{target}] \times[r_{\min },r_{\max }]$,
where $l_t$ is the number of partition regions and may vary over time. 
These partition regions could be regarded as leaves in the decision tree.
The MAB treats the problem as a finite-arm bandit problem with respect to the partition regions, and selects the partition region that maximizes the upper bound of a confidence interval for the expected reward.
After determining the pruning ratio $p_i^t$ and LoRA rank $r_i^t$ for each device $i$, the chosen partition region is split further into multiple regions to form the new partition $S^{t+1}$, allowing the decision tree to grow adaptively as the algorithm proceeds.
S-UCB halts the extension of the decision tree when the leaf diameters (region diameters) fall below a threshold $\delta$, which defines the granularity of pruning ratio and LoRA rank exploration.
Upon receiving all LoRA modules from devices, the server can obtain the completion time $T_i^t$ of device $i$ ($\in[1, N]$).
Then, the agent observes a reward from its interactive environment, which has a crucial impact on future decisions. 
The reward in round $t$ can be defined as follows: 
\begin{equation}
    R (p_i^t, r_i^t)=\frac{\Delta F_i^t \cdot \Delta p \cdot I(\mathbf{B}_i^{\text{total}}) }{\Delta T_i^t }  \label{eq: reward}
    \vspace{-1.5mm}
\end{equation}
where $\Delta F_i^t$ indicates the decrease in local loss during round $t$ and 
$\Delta T_i^t = |T_i^t-\frac{1}{N} \sum_{i^{\prime}=1}^N T_{i^{\prime}}^t|$ represents the gap between the completion time of device $i$ and the average completion time.
A smaller gap implies that the chosen pruning ratio and LoRA rank better align with the device's capabilities, resulting in a higher reward. 
\method aims to determine proper LoRA rank for device $i$ to obtain the LoRA module with  higher importance, \ie, $I(\mathbf{B}_i^{\text{total}})$. 
The pruning granularity in round $t$ is denoted by $\Delta p = p_i^t - p_i^{t-1}$, then a larger granularity is more likely to lead to a higher final pruning ratio, thereby yielding greater inference gains.


Thus, S-UCB aims to maximize the total received rewards by carefully balancing exploration and exploitation. 
It employs an exploration-exploitation strategy to select the most suitable arm from the partitioned space, and seeks to balance exploiting arms that have performed well in the past with exploring arms that may yield higher rewards in the future.
The exploitation and exploration are defined as follows:

(1)\textbf{Exploitation.} Let $N_t(\lambda, S_j^t)=\sum_{s=1}^{t-1} \lambda^{t-s} \mathbbm{1}_{\{(p_i^s, r_i^s) \in S_j^t\}}$ record the number of times that the partition region $S_j^t$ is chosen, where $\lambda \in(0,1)$ is a discount factor. The indicator function $\mathbbm{1}_{\{(p_i^s, r_i^s) \in S_j^t\}}$ is 1 when the action $(p_i^s, r_i^s) \in S_j^t$ and 0 otherwise. The discounted empirical average is given by
\vspace{-1mm}
\begin{equation}
\bar{R}_t(\lambda, S_j^t)=\frac{1}{N_t(\lambda, S_j^t)} \sum_{s=1}^{t-1} \lambda^{t-s} R(p_i^s, r_i^s) \mathbbm{1}_{\{(p_i^s, r_i^s) \in S_j^t\}}
\end{equation}

(2) \textbf{Exploration.} If the agent always selects the pruning ratio and LoRA rank from the partition it currently considers the best, it may overlook another partition with a higher expected reward. 
To address this, S-UCB incorporates an exploration term into the upper bound.
Let $n_t(\lambda)=\sum_{j=1}^{l_t} N_t(\lambda, S_j^t)$ hold and the discounted padding function is defined as:
\vspace{-1mm}
\begin{equation}
c_t(\lambda, S_j^t)=\sqrt{\frac{2 \log n_t(\lambda)}{N_t(\lambda, S_j^t)}}
\vspace{-1mm}
\end{equation}
The upper confidence bound in S-UCB is defined as: 
\begin{equation}
U_t(S_j^t)=\bar{R}_t(\lambda, S_j^t)+c_t(\lambda, S_j^t)
\end{equation}
The region $S_j^t$ with the largest $U_t$ will be chosen. 
To achieve this, S-UCB adaptively learns the partitions and determines the pruning ratios and LoRA ranks for all devices without any prior knowledge. 
The performance of the arm-pulling policy is evaluated by regret, defined as the difference between the expected reward from selecting the optimal arms $(p_i^*, r_i^*)$ and the reward obtained by the given policy. 
The goal of S-UCB is to minimize the cumulative regret over 
$T$ rounds.
\vspace{-2mm}
\begin{equation}
\min \sum_{t=1}^T  \mathbb{E} (R(p_i^*, r_i^*)-R(p_i^t, r_i^t)) 
\vspace{-2mm}
\end{equation}

\subsection{Model Pruning}  \label{sec: model pruning}
To enhance the performance of LLM fine-tuning with heterogeneous data, \method employs a loss-based pruning strategy that adaptively prunes 
redundant LLM parameters for various downstream tasks.
Given that the importance of LLM parameters varies across different downstream tasks, 
\method needs to dynamically determine pruning ratios for heterogeneous devices, which will be elaborated in \S \ref{sec:evaluation}. 
Once the specific pruning ratio $p_i^t$ is determined for device $i$ in round $t$, the device performs model pruning on its local frozen weight $\mathbf{W}_i^0$.
For  structured pruning of LLMs, the goal is to remove the weights  with minimal impact on the model's predictions, as indicated by their effect on the loss function.
Let $\mathbf{W}_i^{m, n} \in \mathbf{W}_i^0$ on device $i$ represent the weight connecting the $m$-th input to the $n$-th output in local model's weight $\mathbf{W}_i^0$. 
Then, the importance of $\mathbf{W}_i^{m, n}$ can be quantified by measuring its impact on the loss after removal.
The induced error of $ \mathbf{W}_i^{m, n}$ can be denoted as
\vspace{-1mm}
\begin{equation}
I(\mathbf{W}_i^{m,n})=\left( F_i(\mathbf{W}_i^0)-F_i(\mathbf{W}_i^0 | \mathbf{W}_i^{m,n}=0)\right)^2
\vspace{-1mm}
\end{equation}
Computing $I(\mathbf{W}_i^{m,n})$ is computationally
expensive.
Hence, the first-order Taylor expansion could be used to approximate the importance of $I_i^{m,n}$ following \cite{molchanov2019importance}:
\vspace{-3mm}
\begin{equation}
    \hat{I}(\mathbf{W}_i^{m,n})=(\mathbf{W}_i^{m, n} \cdot \frac{\partial F_i}{\partial \mathbf{W}_i^{m, n}} )^2
    \vspace{-2mm}
\end{equation}

However, in PEFT, gradients of frozen weights are not directly accessible and only the gradients of LoRA weights are computed during model fine-tuning. 
APT\cite{zhao2024apt} evaluates weight importance by calculating the product of activations and their gradients, which demands substantial computational resources and memory. 
To this end, a LoRA-guided criterion \cite{zhang2023loraprune} that leverages the gradients of LoRA is introduced to evaluate the importance of LLM weights effectively. 
By setting the weight $(\mathbf{B A})_i^{m, n}=-\mathbf{W}_i^{m, n }$ when the weight $\mathbf{W}_i^{m, n} \in \mathbf{W}_i^0$  is removed, we can estimate the importance of $\mathbf{W}_i^{m, n}$ using following gradient-based manner:
\vspace{-1mm}
\begin{align} \label{eq:lora_prune}
    \hat{I}_i^{m, n} (\mathbf{W}_i^{m,n})= & [(\frac{\partial F_i}{\partial \mathbf{B}_i^{m,:}} \mathbf{A}_i^{:, n}+\mathbf{B}_i^{m,:} \frac{\partial F_i}{\partial \mathbf{A}_i^{:, n }}-\frac{\partial F_i}{\partial \mathbf{B}_i^{m,:}} \frac{\partial F_i}{\partial \mathbf{A}_i^{:, n}}) \cdot \notag\\
     & (\mathbf{W}_i^{m, n}+(\mathbf{B} \mathbf{A})_i^{m ,n})]^2  
\end{align}

This pruning criterion only requires the computation of gradients for $\mathbf{A}$ and $\mathbf{B}$ with the approximation in Eq. (\ref{eq:lora_prune}), effectively circumventing the issue of inaccessibility of gradients for the frozen LLM. 
Drawing inspiration from previous work \cite{ma2023llm, fang2023depgraph}, \method prunes heads for the Attention layer and channels for the FNN layer, respectively.
However, in structured pruning, it is essential to consider that pruned weights may exhibit dependencies with other weights due to their interconnected nature. 
Specifically, the weight dependency rule in \method follows the approach in \cite{ma2023llm, zhang2023adalora}.
Let the set of linear layers $\Psi = \{query, key, value, out\}$ and $\Phi = \{ffn1, ffn2\}$ (or $\Phi = \{gate, up, down\}$ for LLaMA models) denote the weight dependency for Attention and FNN layers, respectively.
We treat the connected weights within each model group as a unit, ensuring that all weights within the same group are pruned simultaneously. 
Note that pruning partial weights in a group is avoided as it may lead to misaligned intermediate representations\cite{ma2023llm}.
Hence, we estimate the group importance by summing the importance of all weights within the same group.
Then, the importance of the $g$-th ($ g \in \{1,  \ldots, G\}$) group of the $l$-th layer can be formally denoted as:
\vspace{-2mm}
\begin{equation} \label{eq:group importance}
\hat{\Theta}_i^{l,g}=\sum_{\mathbf{W}_i^{m, n} \in \mathbb{G}} \hat{I}(\mathbf{W}_i^{m,n})
\end{equation}
where $\hat{\Theta}_i^{l}  \in \mathbb{R}^{1 \times G}$ represents the importance of groups, $\mathbb{G}$ denotes a set of connected weights within a group and $G$ is the number of candidate groups in the $l$-th layer.

After several fine-tuning iterations (\eg, $\tau$ = 20 in our experiment), we estimate the weight importance for each group by calculating the moving average with $\eta \in [0,1]$ as
\vspace{-1mm}
\begin{equation}
\Theta_{i,k}^{l,g} = \eta \Theta_{i,k-1}^{l,g} + (1 - \eta) \hat{\Theta}_{i,k}^{l,g} 
\vspace{-1.5mm}
\end{equation}
where $\Theta_{i,k}^{l,g}$ denotes the group importance scores at the $k$-th iteration. 
After repeating $\tau$ iterations, we could obtain $\Theta_{i,\tau}^{l,g}$ and then prune the unimportant groups based on the derived pruning ratio from our MAB-based algorithm in \S \ref{subsec: MAB}. 
Specifically, each device applies a binary mask $M_i^l \in\{0,1\}^{1 \times G}$ for each pruned layer, and then selects the top $p_i^t\%$ unimportant groups and sets the corresponding binary masks $M_i^{l,g}$ to 0, while others are as 1. 
Regarding the binary masks, the forward propagation for each pruned layer can be denoted as $\mathbf{z}=(\mathbf{W}_i^0 x +\mathbf{ B A }x) \odot M$, where $\odot$ denotes the Hadamard product between the masks and their corresponding matrices. 
\vspace{-2mm}

\subsection{Model Fine-tuning}  \label{sec: model fine-tuning}
\method is designed to dynamically assign LoRA ranks for heterogeneous devices during fine-tuning to enhance model performance.
To balance the performance gains against the additional computational and communication costs, we adaptively assign larger LoRA ranks for stronger devices to improve model performance, while weaker devices receive smaller LoRA ranks to mitigate the straggler effect.
As a result, we recover the performance of the pruned LLM on each device without slowing down model convergence.
To determine which devices should be allocated the larger LoRA ranks, \method first calculates the importance of each device's LoRA modules.
Specifically, the importance of each LoRA module is represented as the summation of the parameter importance within matrix $\mathbf{B}$, \ie, $I(\mathbf{B}_i)=\sum_{m, n} I(\mathbf{B}_{i}^{m, n})$\footnote{The importance scores calculated using $\mathbf{B}_i$ and $\mathbf{A}_i$ are equal.}. 
We add LoRA modules into the linear layer $ \Omega=\{\Psi, \Phi\}$ within each LLM transformer layer. 
Thus, the importance of LoRA modules within the $l$-th layer is calculated as follows:
\begin{equation}
    I(\mathbf{B}_i^l)=\sum_{\sigma \in \Omega} \sum_{m, n} I(\mathbf{B}_{i,\sigma}^{m, n} )
    \vspace{-1mm}
\end{equation}

To enhance the evaluation of LoRA module importance, we incorporate singular values into the assessment of the matrix $\mathbf{B}$. 
Larger singular values indicate that the associated weights capture more critical information, thus we characterize the importance of weights using the average singular value of each trainable weight.
The overall importance of the LoRA module for device $i$ can be formulated as:
\vspace{-1mm}
\begin{equation}
I(\mathbf{B}_i^{\text{total}})=\sum_l I(\mathbf{B}_i^l) + \frac{1}{d_1}  \sum_{s=1}^{d_1} \sum_{e=1}^{d_2} \lambda_{i,s}^e
\vspace{-1mm}
\end{equation}
where $\lambda_{i, s}^e$ denotes the $e$-th singular value of the $s$-th LoRA module on device $i$, with $d_1$ and $d_2$ representing the total number of LoRA modules and singular values in the $s$-th LoRA module, respectively. 
Efficient fine-tuning is achieved by increasing the LoRA rank of devices with higher $I(\mathbf{B}_i^{\text{total}})$ values. 
The LoRA rank $r_i^t$ is determined using a MAB-based algorithm (see \S \ref{subsec: MAB}).
To ensure fine-tuning stability when increasing LoRA ranks and converting $\mathbf{B}_i^t  \in \mathbb{R}^{d \times r_i^t}$, $\mathbf{A}_i^t \in \mathbb{R}^{r_i^{t} \times k}$ to $\mathbf{B}_i^{t+1} \in \mathbb{R}^{d \times r_i^{t+1}}$, $ \mathbf{A}_i^{t+1} \in \mathbb{R}^{r_i^{t+1} \times k}$, we  concatenate zeros in $\mathbf{B}_i^t$ and random Gaussian initialized parameters $N(0, \sigma^2)$ in $\mathbf{A}_i^t$, following the LoRA initialization method.
This ensures that the layer’s output
remains unchanged before and after new parameters are added. 
When smaller LoRA ranks are adopted for device $i$, the downloaded LoRA weights  $\mathbf{A}_i^t$ and $\mathbf{B}_i^t$ are shrunk using $\Lambda(\mathbf{A}_i^{t+1}, \mathbf{B}_i^{t+1}, r_i^{t+1})$, which reduces the LoRA weights to the smaller rank $r_i^{t+1}$. 
\vspace{-2mm}
\subsection{LoRA Aggregation} \label{sec: LoRA aggregation}
\vspace{-1mm}
\method allows each device to upload LoRA modules $\mathbf{B}$ and $\mathbf{A}$ instead of the full model during the fine-tuning, which reduces the communication and aggregation overhead.
However, the heterogeneous LoRA modules with different ranks from all devices cannot be aggregated directly.  
A straightforward way is using zero-padding to all the received LoRA modules with $r_i^{t}<\max_{i} \{r_i^{t}\}$ and then performing simple averaging over the modules. 
However, such aggregation has been shown to bias the model toward higher-rank devices \cite{cho2023heterogeneous}.
Therefore, before aggregation, \method first reconstructs these LoRA modules to the full model as 
\vspace{-1mm}
\begin{equation}
    \Delta \mathbf{W}_i^{t}=\mathbf{B}_i^{t}  \cdot \mathbf{A}_i^{t}
    \vspace{-1.5mm}
\end{equation}

Then, the server sums the LoRA modules with aggregation weights $c_i^t$ which are proportional to the importance of devices' LoRA modules, \ie,
\vspace{-1mm}
\begin{equation}
    \Delta \mathbf{W}_i^{t+1}  =  \frac{1}{N}\sum_{i=1}^{N} c_i^{t} \cdot \Delta \mathbf{W}_i^{t}
    \vspace{-2mm}
\end{equation}
where $c_i^t = I(\mathbf{B}_i^{\text{total}})/\sum_{i \in N} I(\mathbf{B}_{i}^{\text{total}})$. 
The updated global LoRA modules are stored in the server and will be distributed to participating devices for further fine-tuning, or be combined with pruned LLMs to perform inference for various downstream tasks.   
Alongside the LoRA aggregation process, the server should utilize the collected state information to determine the refining configurations for different devices in the next round, aiming to balance efficiency and accuracy.

\section{Implementation and Evaluation}\label{sec:evaluation}
\subsection{System Implementation} 
We have implemented \method prototype with about 3k lines of custom code using FedPETuning \cite{lai2022fedscale}, an open-source platform for federated fine-tuning.
The prototype adopts a server-device architecture tailored for LLM deployment on resource-constrained and heterogeneous devices, including Jetson TX2, NX, and AGX.
The software platform is built on Docker Swarm \cite{naik2016building}) and Pytorch \cite{paszke2019pytorch}.
Docker Swarm, a distributed software development kit, facilitates building distributed systems and provides real-time monitoring of each device’s operational status.
PyTorch supports the implementation of model fine-tuning on devices, offering platform independence while leveraging platform-specific backend acceleration.
To accelerate on-device fine-tuning, we leverage NVIDIA-provided development packages\footnote{https://forums.developer.nvidia.com/t/pytorch-for-jetson/72048} to fully exploit the hardware capabilities of Jetson devices. 
Communication between the server and devices is streamlined through MPI (Message Passing Interface)  \cite{gabriel2004open}, which offers efficient data exchange with functions such as \textit{comm.send(data, dest, tag)} and \textit{comm.recv(source, tag)}.

\setlength{\parindent}{0pt}
\textbf{Hardware.}\hspace{0.7em}
\setlength{\parindent}{10pt}
For the implemented prototype system, we employ an AMAX deep learning workstation, which is equipped with an Intel(R) Xeon(R) Platinum 8358P CPU, 8 NVIDIA GeForce RTX A6000 GPUs (each with 48GB memory), and 512 GB of RAM, as the server.
Besides, we incorporate 80 NVIDIA commercial developer kits as devices, including 30 Jetson TX2 kits, 40 Jetson NX kits, and 10 Jetson AGX kits, to establish a heterogeneous system. 
The detailed technical specifications of these devices are listed in Table \ref{table: jetson}. 




\setlength{\parindent}{0pt}
\textbf{Simulation of System Heterogeneity.}\hspace{0.7em}
\setlength{\parindent}{10pt}
To simulate the heterogeneous computing and communication capabilities among devices, the prototype system is configured as follows:
\begin{itemize}[leftmargin=0.5cm]

\item \textit{For Computation.} These devices (\ie, Jetson TX2, NX, and AGX) are  configured to work with different modes, where the number of active CPU cores and CPU/GPU frequency levels can be adjusted to work with varying computing capabilities. 
Specifically, TX2 supports four configurable modes, while NX and AGX offer eight modes. 
Devices working in different modes exhibit diverse capabilities. 
For example, compared to the TX2 with mode 1 (\ie, the lowest performance mode), the AGX with mode 0 (\ie, the highest performance mode) can shorten the training time by 100$\times$.
To further reflect the time-varying on-device resources, each mode of device is randomly adjusted every 20 rounds.

\begin{table}[t]
    \caption{Technical specifications of end devices.}
    \label{table: jetson}
    \centering
    \scalebox{0.87}{
      \begin{tabular}{c|c|c|c}
        \hline
         & \textbf{AI Performance} & \textbf{GPU} & \textbf{ROM} \\ \hline
        TX2 & 1.33 TFLOPS & 256-core Pascal & 8 GB LPDDR4 \\ \hline
         NX  & 21 TOPS & 384-core Volta & 8 GB LPDDR4x\\ \hline 
         AGX & 22 TOPS & 512-core Volta & 32 GB LPDDR4x \\  \hline 
      \end{tabular}}
      \vspace{-0.3cm}
\end{table}

\item \textit{For Communication.}
All devices are connected to the server via 2.4 GHz WiFi routers and are randomly shuffled into four groups placed at varying distances, \ie, 2m, 8m, 14m, and 20m far away from the WiFi routers in the prototype system. 
Consequently, the communication bandwidth between the server and each device fluctuates dynamically, influenced by random channel noise and inter-device interference.  
The bandwidth between each device and the server is measured by iperf3 \cite{tirumala1999iperf}, which varies between 1Mb/s and 30Mb/s during training.
\end{itemize}

\subsection{Experimental Setup} 

\setlength{\parindent}{0pt}
\textbf{Datasets and Models.}\hspace{0.7em}
We evaluate the performance of \method on real-world datasets across three applications, with the statistics of each dataset summarized in Table \ref{tab: datasets}:
\begin{itemize}[leftmargin=0.5cm]
\item \textit{Sentiment Analysis}: We adopt Stanford Sentiment Treebank (SST-2) \cite{wang2018glue} dataset,  which 
consists of movie reviews annotated with human sentiments. 
For this task, we employ the RoBERTa model \cite{liu2019roberta} with 110M parameters. 
\item \textit{Textual Entailment}: The Multi-Genre Natural Language Inference (MNLI) dataset from the GLUE benchmark \cite{wang2018glue} is adopted. 
The task is to predict whether one sentence can be inferred from another.
The model employed on MNLI is the same as that on SST-2,  \ie, RoBERTa.
\item \textit{Question Answering}: LLaMA-7B is fine-tuned on dolly-15K \cite{conover2023free}, partitioned across 8 heterogeneous AGX devices based on the categories of questions. We evaluate its performance with four tasks on HELM \cite{liang2022holistic}. 
Recently, the industry has focused on deploying models with approximately 8 billion parameters on terminal devices \cite{xiao2024large, dhar2024empirical}.
\end{itemize}

\begin{table}[!t]
\caption{Statistics of datasets in experiments.}
\vspace{0.1cm}
\label{table: datasets}
\centering
\scalebox{0.9}{
    \begin{tabular}{ccccc}
    \hline
    Dataset & Training samples & Testing samples & Metric   \\ 
    \hline 
    SST-2 \cite{wang2018glue} & 67,349 & 1,821 & Accuracy  \\ 
    MNLI \cite{wang2018glue} & 392,702 & 9,815 & Accuracy \\ 
    Dolly \cite{conover2023free} &  15,015 & NA & Accuracy \\ \hline
    
    \end{tabular}
} \label{tab: datasets}
\vspace{-0.3cm}
\end{table}


\begin{table*}[ht]
\centering
\caption{The LLaMA-7B model performance of \method and baselines under 30\% sparsity.}
\scalebox{0.80}{
\begin{tabular}{l|ccccc|cccc}
\toprule
 \textbf{Method} & \textbf{HellaSwag} & \textbf{MMLU} & \textbf{BoolQ} & \textbf{OBQA} & \textbf{Avg. (↑)} & \textbf{Train. Time (↓)} & \textbf{Train. Mem. (↓)} & \textbf{Inf. Time (↓)} & \textbf{Inf. Mem. (↓)} \\

\midrule

     HETLoRA & 77.1 & 43.2 & 76.8 & 56.5  & 63.4 &  100\% & 100\% & 100\% & 100\% \\ 
\midrule

     FedLoRA + Prune     & 67.9    & 25.7  & 65.0 & 41.8  & 50.1  & 176.2\%  & 95.5\%      & 84.7\% & 62.5\%\\
     LLMPruner + FedLoRA & 68.2    & 24.9  & 65.6 & 41.3  & 50.0  & 159.4\%  & 95.5\%     & 83.4\% & 67.9\% \\
     FedAPT              & 69.3    & 30.8  & 66.4 & 42.6  & 52.3  & 252.6\%  & 127.1\% 
     & 87.9\% & 64.1\% \\
     FedSpine (Ours)     & \textbf{71.2}  & \textbf{36.6} & \textbf{67.5} & \textbf{42.7} & \textbf{54.5} & \textbf{97.3\%} & \textbf{81.6\%} & \textbf{85.3\%} & \textbf{63.2\%}\\
    
\bottomrule
\end{tabular} \label{tab: performance of LLaMA}
}
\vspace{-3mm}
\end{table*}

\textbf{Setting of Data Heterogeneity.}\hspace{0.7em}
To simulate various degrees of statistical heterogeneity, we partition the datasets by using the Dirichlet distribution as the class priors \cite{hsu2019measuring}.
In particular, we sample $D \sim \operatorname{Dir}(\alpha)$ and allocate data $D_i$ to device $i$, where $\alpha$ determines the degree of non-IID. 
A smaller $\alpha$ generates a high label distribution shift, while a larger $\alpha$ indicates more balanced distributions.
As $\alpha \rightarrow $ $\infty$, all devices have identical distributions (\ie, IID); as $\alpha \rightarrow$ 0, each device holds data samples from only one class, which indicates a high degree of statistical heterogeneity.  
We specify 4 values for $\alpha$ (10, 1.0, 0.5, 0.1) to generate diverse data distributions. 
Unless otherwise specified, our experiment uses a default Dirichlet parameter setting of $\alpha=0.5$. 

\textbf{Baselines.}\hspace{0.5em}
We measure the effectiveness of \method through comparing it with four approaches.
(\romannumeral1) HETLoRA \cite{cho2023heterogeneous}. This method allows devices to fine-tune different local models with heterogeneous LoRA layers to deal with device heterogeneity. 
(\romannumeral2) FedLoRA \cite{zhang2023fedpetuning} + Prune \cite{kwon2022fast}. FedLoRA is a foundational method to federated LoRA-based fine-tuning, as introduced in \cite{zhang2023fedpetuning}. We pair it with Mask Tuning, a SOTA post-training structured pruning method based on fisher information. 
(\romannumeral3) LLMPruner \cite{ma2023llm} + FedLoRA. LLMPruner is the  state-of-the-art task-agnostic pruning method for LLMs that prunes its blocks or channels based on salience metrics, while using LoRA for fast performance recovery.
We compare \method to LLMPruner combined with the FedLoRA method on the same data to ensure a fair comparison.
(\romannumeral4) FedAPT. APT\cite{zhao2024apt} adaptively identifies LLMs’ pruning and tuning parameters during fine-tuning. Although we apply APT in a federated setting (see \S \ref{sec: combination of pruning and fine-tuning}), it fails to account for device heterogeneity. 
For fair comparisons, we implement the above pruning baselines with iterative  fine-tuning and pruning, where the pruning process for all methods is progressive.

\textbf{Metrics.}\hspace{0.7em}
The following metrics are adopted to evaluate the performance of \method and baselines. 
\begin{itemize}[leftmargin=0.5cm]
\item \textit{Test Accuracy}: We measure the accuracy of models trained by different approaches on the test datasets, defined as the proportion of correctly predicted data. 
    Specifically, we evaluate the average test accuracy of the local model (a combination of the local LoRA module and LLMs) on devices in each round, and record the final test accuracy across all methods.
\item \textit{Training Metrics}: We record training peak memory and the total wall clock time required to fine-tune a model to achieve a target accuracy.
    For a fair comparison, the target accuracy is set as the achievable accuracy by all methods.
\item \textit{Inference Metrics}: We report the inference time and peak memory for each device on their respective datasets. 
\end{itemize}

\begin{table}[!t]
\vspace{-3mm}
\caption{The RoBERTa model performance of \method and baselines under 50\% sparsity.}
\vspace{0.1cm}
\label{table: datasets}
\centering
\scalebox{0.7}{\begin{tabular}{c|cc|cccc}
\hline Methods & \makecell[c]{SST-2}  & \makecell[c]{MNLI} & \makecell[c]{Train. \\ Time (↓)} & \makecell[c]{Train. \\ Mem. (↓)} & \makecell[c]{Inf. \\ Time (↓)} & \makecell[c]{Inf. \\ Mem. (↓)}  \\
\hline 
HETLoRA      & 96.6 & 85.7 & 100.0\% & 100.0\% & 100.0\% & 100.0\% \\
Fed. + Prune & 95.7 & 84.9 & 285.7\% & 89.3\% & 49.6\%  & 78.8\% \\
LLM.+ Fed.   & 95.5 & 83.3 & 364.3\% & 89.3\% & 49.1\%  & 79.2\% \\
FedAPT       & 96.0 & 84.3 & 492.9\% & 132.7\% & 52.3\%  & 83.1\% \\
FedSpine     & 96.4 & 86.0 & 71.4\% & 114.5\% & 51.4\%    & 81.6\% \\
\hline
\end{tabular}} \label{tab: performance of RoBERTa}
\vspace{-0.3cm}
\end{table}

\textbf{Critical Hyper-parameters.}\hspace{0.7em}
By default, all experiments will run 100 rounds to ensure that the target pruning ratio is achieved, and pruning will stop when the target pruning ratio is reached on the device.
We use a mini-batch size of 32, 16, 10 for SST-2, MNLI, and dolly. The learning rate is set as 0.0005 \cite{zhang2023fedpetuning, yao2024ferrari}.
The ranks of FedLoRA are set to 32, while the ranks of HETLoRA are set within \{2, 4, 8, 16, 32\} 
\cite{cho2023heterogeneous}. 
The inference test batch size is 128 for RoBERTa and 32 for LLaMA-7B \cite{zhao2024apt}, respectively. 
All pruning baselines and \method have the same local fine-tuning steps, \ie, 20.

\subsection{Numerical Results}
\textbf{Overall performance.}\hspace{0.7em} 
We investigate the performance of \method and baselines  when deployed across heterogeneous devices.
The results in Table \ref{tab: performance of RoBERTa} show that \method always achieves comparable accuracy to HETLoRA on RoBERTa, 
demonstrating that its adaptive pruning ratios and LoRA ranks do not compromise model performance.
As shown in Table \ref{tab: performance of LLaMA}, for LLaMA-7B models with 70\% of parameters retained, \method  recovers 86.0\% model performance on average  while using only 81.6\% the training memory compared to HETLoRA.
Additionally, \method significantly reduces training time, achieving a speedup of 1.4$\times$-6.9$\times$ over baselines. 

\begin{figure}[t]
\centering
\subfigure[SST-2]{
    \includegraphics[width=0.23\textwidth,height=3.5cm]{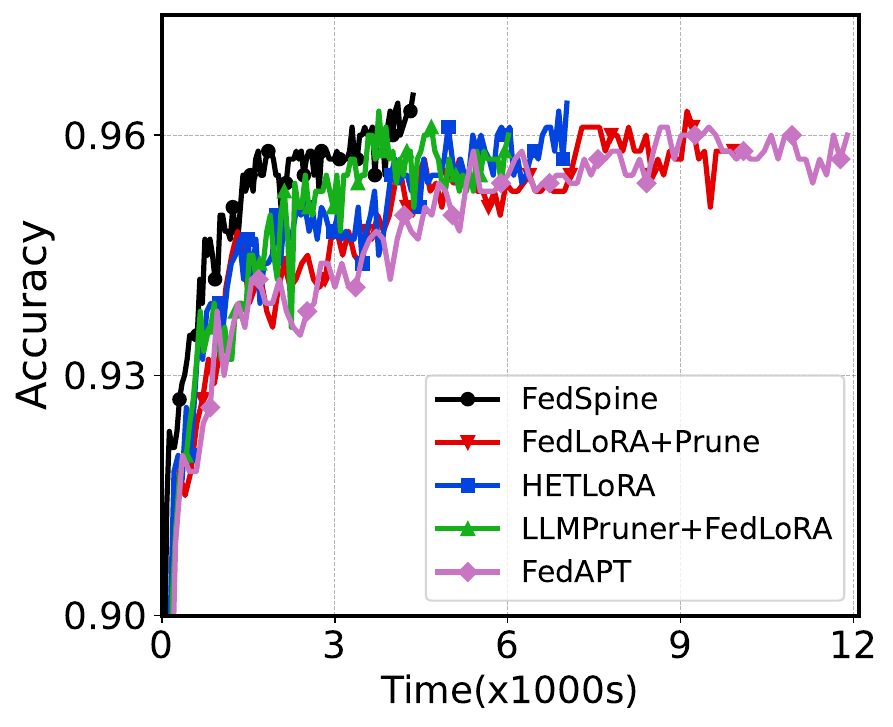}\label{fig: SST-2_acc}}
\subfigure[MNLI]{
    \includegraphics[width=0.23\textwidth,height=3.5cm]{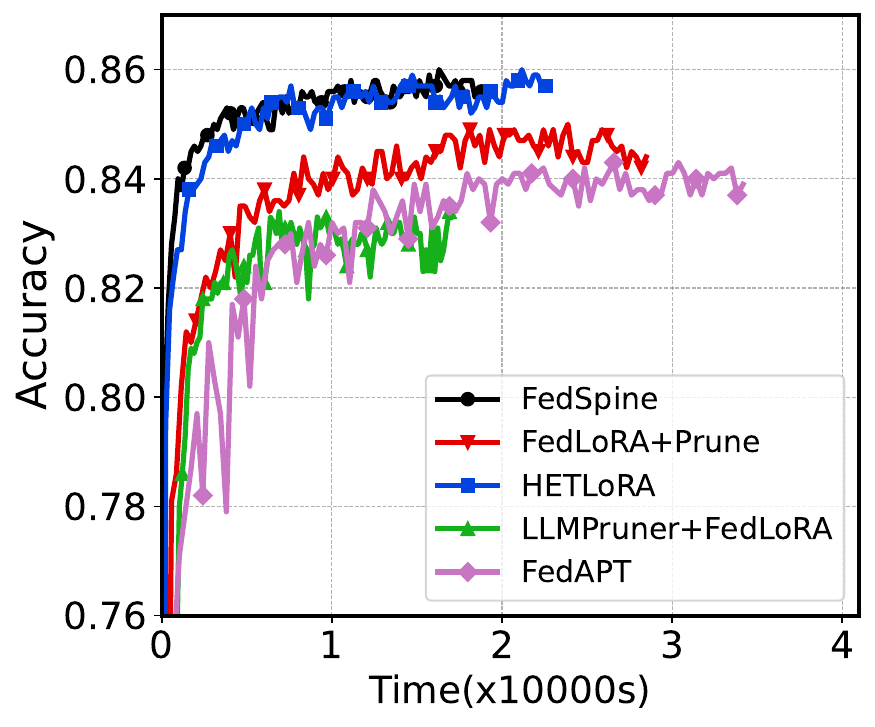}\label{fig: MNLI_acc}}
\caption{Performance comparison of RoBERTa on SST-2 and MNLI.} 
\label{fig: Performance of RoBERTa}
\vspace{-2mm}
\end{figure}

Figure \ref{fig: Performance of RoBERTa} illustrates the test accuracy with time passed on SST-2 and MNLI and Table \ref{tab: performance of LLaMA} shows the performance of different methods for fine-tuning LLaMA-7B.  
From these results, we derive three major observations.
Firstly, \method outperforms HETLoRA in training speed and requires substantially less time to achieve comparable accuracy.
For instance, \method takes 1150.6s to achieve 83\% accuracy for RoBERTa on MNLI, while HETLoRA takes 1610.8s. 
\method provides 1.4$\times$ speedup compared to HETLoRA. 
While HETLoRA allows heterogeneous devices to hold different LoRA ranks, it does not incorporate any parameter reduction techniques and the performance gap of heterogeneous devices is not entirely eliminated.
Furthermore, \method demonstrates superior inference performance compared to HETLoRA due to its reduction in model parameters.

Secondly, \method converges much faster than FedAPT, 
and achieves a speedup of nearly 4.1$\times$ for RoBERTa on SST-2. 
This is because FedAPT accelerates fine-tuning with uniform LLM pruning and LoRA rank assignments across devices, which fail to account for their heterogeneity. 
Since the pruned models and assigned LoRA ranks are the same size, weaker devices may delay global aggregation.
Unfortunately, uniform pruning can even affect end-task performance, as it  fails to consider the sensitivity of downstream tasks to model parameters, leading to over-pruning or under-pruning on certain devices.
For example, \method achieves 1.7\% higher accuracy than FedAPT on MNLI.
In contrast, \method dynamically prunes local models and adjusts LoRA ranks according to the heterogeneity of devices.
These customized pruning and LoRA module adjustments, aligned with each device’s capabilities, lead to a substantial acceleration in LLM fine-tuning.

Thirdly, \method consistently demonstrates significant advantages over other pruning baselines (\ie, FedLoRA+Prune and LLMPruner+FedLoRA) in both training speed and accuracy performance.
Specifically, \method outperforms over FedLoRA+Prune, providing a 4.0$\times$ speedup and achieving a 1.1\% accuracy improvement on MNLI. 
Compared with LLMPruner+FedLoRA, \method achieves about 5.1$\times$ 
speedup and  2.7\% accuracy improvement on MNLI, correspondingly. 
While both methods allow iterative structured pruning and fine-tuning in our setting, they fail to account for the device heterogeneity. 
Moreover, the lack of coordination between pruning and fine-tuning results in suboptimal performance and poor fine-tuning efficiency. 
In contrast, \method employs adaptive pruning and fine-tuning strategies to achieve superior performance.
It dynamically assigns smaller pruning ratios and lower LoRA ranks to slower devices, reducing local training time, while  allocating larger pruning ratios and higher LoRA ranks to faster devices to preserve model performance.
Consequently, \method maintains higher performance while improving fine-tuning efficiency.

\begin{figure}[t]
\centering
\subfigure[RoBERTa on SST-2]{
    \includegraphics[width=0.23\textwidth,height=3.55cm]{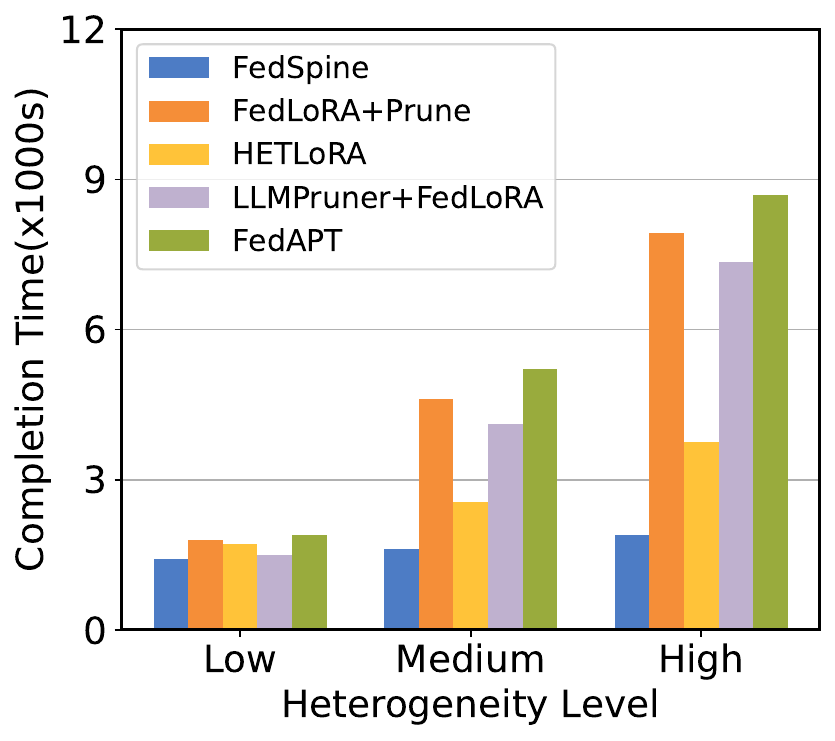}\label{fig: Time_heterogeneity_SST-2}}
\subfigure[RoBERTa on MNLI]{
    \includegraphics[width=0.23\textwidth,height=3.5cm]{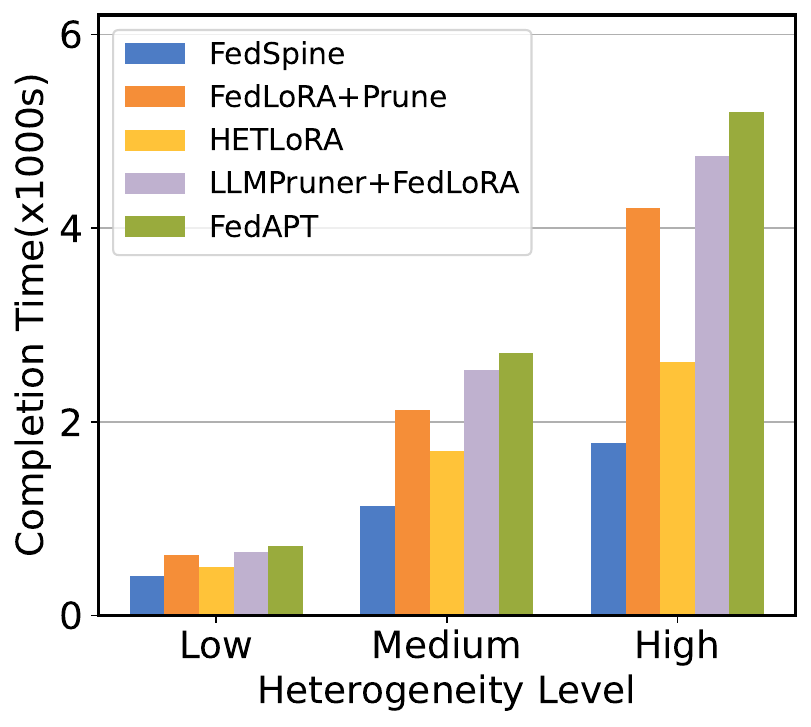}\label{fig: Time_heterogeneity_MNLI}}
\caption{Completion time under different heterogeneous levels.} 
\label{fig: Completion time under different heterogeneous levels.}
\vspace{-2mm}
\end{figure}

\begin{figure}[t]
\centering
\subfigure[RoBERTa on SST-2]{
    \includegraphics[width=0.23\textwidth,height=3.5cm]{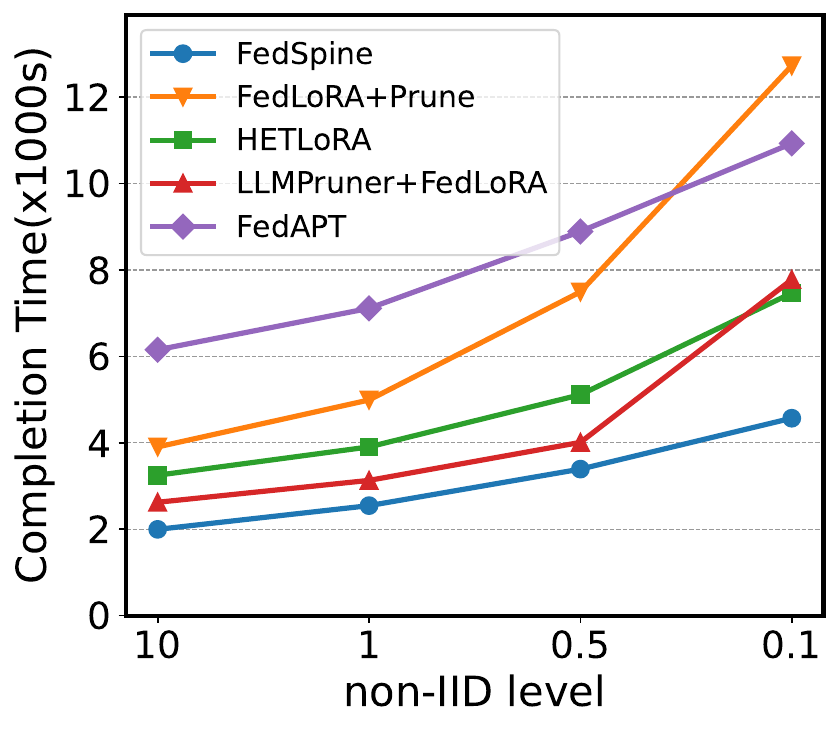}\label{fig: SST-2_noniid}}
\subfigure[RoBERTa on MNLI]{
    \includegraphics[width=0.23\textwidth,height=3.5cm]{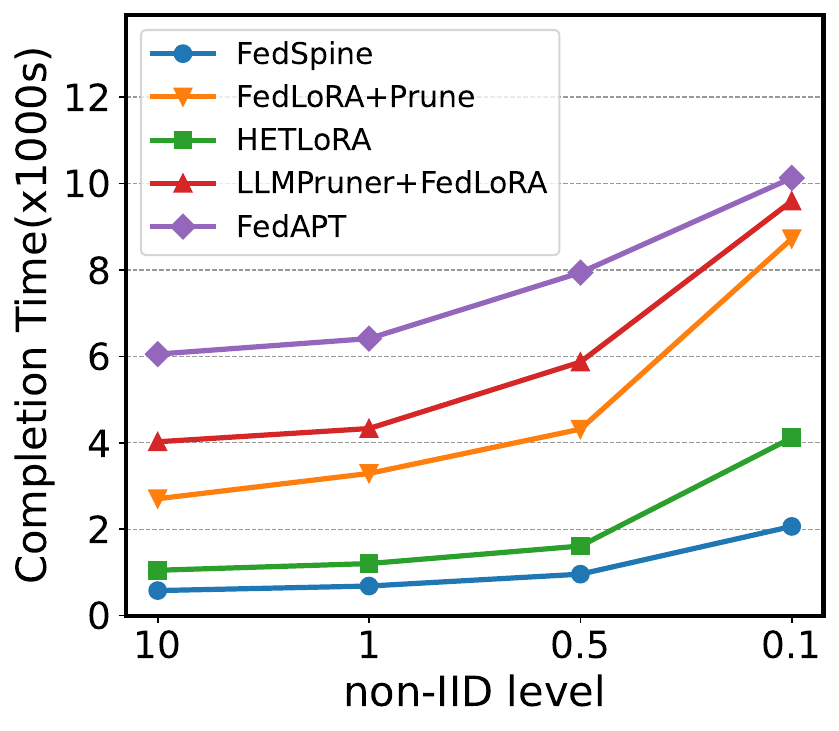}\label{fig: MNLI_noniid}}
\caption{Completion time under different non-IID levels.} 
\label{fig: Completion time under different non-IID levels.}
\vspace{-2mm}
\end{figure}

\setlength{\parindent}{0pt}



\textbf{Effect of Device Heterogeneity.}\hspace{0.7em}
To further evaluate the performance of \method under heterogeneous scenarios, we deploy \method and baselines across the devices with different heterogeneity levels, \ie, \textit{Low}, \textit{Medium}, \textit{High}.
For \textit{Low}, we select 10 AGX devices with the strongest capabilities (see Table \ref{table: jetson}) placed 2m from the WiFi routers.
For \textit{Medium}, we select 5 AGX devices placed 2m away from the routers, and 5 NX devices placed 8m away.
For \textit{High}, we select 3 AGX devices (2m from the routers), 3 NX devices (8m away), and 4 TX2 devices (14m away).
For our experiments, the desired target accuracy of SST-2 and MNLI is 95\% and 80\%, respectively. 
Figure \ref{fig: Completion time under different heterogeneous levels.} shows the required time to reach the target accuracy under different heterogeneous scenarios.
We observe that, from \textit{Low} to \textit{High}, the required time
to reach the target accuracy increases accordingly. This is
expected because less capable devices are introduced into
the system and devices are further away from WiFi routers.
Nevertheless, \method still takes less time to reach the target accuracy compared with the baselines.
When fine-tuning RoBERTa on MNLI in the heterogeneity level of \textit{High}, \method achieves 2.36$\times$, 1.47$\times$, 2.66$\times$, 2.91$\times$ speedup compared to FedLoRA+Prune, HETLoRA, LLMPruner+FedLoRA, and FedAPT, respectively.
Moreover, the performance gap widens with the increase of heterogeneity level.
For example, \method improves the performance over FedLoRA+Prune by 1.27$\times$ in \textit{Low}, 2.85$\times$ in \textit{Medium}, and 4.2$\times$ in \textit{High} for RoBERTa on SST-2.
This is because \method determines the appropriate pruning ratios and LoRA ranks for the weak devices that are newly introduced into the system, so that each device can still train the model that best suits its capability, leading to a slight increase in completion time.
These results demonstrate that \method is robust for diverse heterogeneous scenarios and can fully utilize the limited resources of devices.

\textbf{Effect of Non-IID Data.}\hspace{0.7em} 
We proceed to investigate how \method performs under varying levels of non-IID data distributions.
Figure \ref{fig: Completion time under different non-IID levels.} illustrates the time required for \method and baselines to reach the target accuracy across different levels of non-IID data ($\alpha$ = 10, 1.0, 0.5, 0.1).
We set the target accuracy of SST-2 and MNLI as 95\% and 83\%. 
From the results, we observe that all approaches suffer from performance degradation with the increase of non-IID levels.
Since the LoRA modules fine-tuned on non-IID data differ across devices, aggregating these divergent LoRA modules may degrade the fine-tuning performance and result in more rounds to achieve convergence.
Despite this, \method still outperforms the baselines across all non-IID levels with the slightest increase in the completion time. 
For instance, even in the non-IID level of $\alpha$ = 0.1, \method can reduce the completion time by 64.1\%, 38.9\%, 41.3\%, and 58.2\% compared to FedLoRA+Prune, HETLoRA, LLMPruner+FedLoRA, and FedAPT for RoBERTa on SST-2. 
These results highlight the effectiveness and robustness of our framework in handling non-IID data distribution.



\textbf{Ablation Study.}\hspace{0.7em}
\method employs two core strategies, \ie, adaptive pruning ($\mathbf{A}_{\mathrm{P}}$) and adaptive fine-tuning ($\mathbf{A}_{\mathrm{T}}$) for heterogeneous devices to enhance its performance. 
To evaluate the impact of these components, we conduct an ablation study by removing $\mathbf{A}_{\mathrm{P}}$ and $\mathbf{A}_{\mathrm{T}}$ and the results are presented in Table \ref{tab: ablation study of RoBERTa} and Table \ref{tab: ablation study of LLaMA}. 
In the first case, we disable $\mathbf{A}_{\mathrm{P}}$ and only fine-tune LLMs with adaptive tuning strategies. 
In such settings, \method can be recognized as FedLoRA with dynamic LoRA ranks for each device.  
Therefore, the inference efficiency of the fine-tuned LLMs is comparable to those fine-tuned with HETLoRA.
Without pruning, the model performance of RoBERTa and LLaMA achieves a higher accuracy than \method.
In addition, we observe that the RoBERTA with \method w/o $\mathbf{A}_{\mathrm{P}}$ requires 11.6\% less training time than \method while costing only 97.7\% of memory.
Meanwhile, the training memory cost of \method without $\mathbf{A}_{\mathrm{P}}$ will be higher than that of FedLoRA+Prune since the total LLM parameters number is fixed, yet the tuning parameter number of \method will grow larger than static FedLoRA tuning.  
In the second case, we disable $\mathbf{A}_{\mathrm{T}}$ where tuning parameters are static when pruning RoBERTa and LLaMA-7B models.  
Without $\mathbf{A}_{\mathrm{T}}$, the model performance decreases to 94.9\% and 84.3\%.
At the same time, $\mathbf{A}_{\mathrm{T}}$ speeds the model converge 34.2\% faster than HETLoRA.
When ablating $\mathbf{A}_{\mathrm{T}}$ in LLaMA model pruning, as shown in Table \ref{tab: ablation study of LLaMA}, we observe that $\mathbf{A}_{\mathrm{T}}$ recovers the model performance under 30\% pruning setting (54.5 compared to 53.5).
we conduct that $\mathbf{A}_{\mathrm{T}}$ substantially improves LLM fine-tuning speed and end-task performance.  

\begin{table}[!t]
\caption{Ablation effect of adaptive pruning.}
\vspace{0.1cm}
\label{table: datasets}
\centering
\scalebox{0.9}{
\begin{tabular}{l|cc|cc}
\hline  Method  & SST-2 &  MNLI &  Train. Time (↓) &  Train. Mem. (↓) \\
\hline  \method  & 96.4 & 86.0 &  71.4\% &  114.5\% \\
 $w/o$  $\mathbf{A}_{\mathrm{P}}$ & 96.3 & 85.8  & 65.8\% &  97.7\% \\
 $w/o$  $\mathbf{A}_{\mathrm{T}}$ & 94.9 & 84.3  &  92.8\% &  104.5\% \\
\hline
\end{tabular}}  \label{tab: ablation study of RoBERTa}
\vspace{-0.3cm}
\end{table}

\begin{table}[!t]
\caption{Ablation effect of adaptive fine-tuning.}
\vspace{0.1cm}
\label{table: datasets}
\centering
\scalebox{0.7}{
\begin{tabular}{l|ccccc|cc}
\hline  Method                    & HellaSwag &  MMLU & BoolQ & QBQA & Avg. (↑) &  Train. Mem. (↓) \\
\hline  \method                   & 71.2      & 36.6  & 67.5  & 42.7 & 54.5 &  81.6\% \\
 $w/o$  $\mathbf{A}_{\mathrm{P}}$ & 78.3      & 47.2  & 76.8  & 58.7 & 65.3 & 97.4\%  \\
 $w/o$  $\mathbf{A}_{\mathrm{T}}$ & 70.4      & 35.8  & 66.3  & 41.4 & 53.5 & 82.3\%  \\
\hline
\end{tabular}}  \label{tab: ablation study of LLaMA}
\vspace{-0.3cm}
\end{table}

\begin{table}[!t]
\caption{Impact of different $\tau$. “Average” denotes the average performance on four classification datasets.}
\label{table: iteration frequency}
\centering
\scalebox{0.7}{
\begin{tabular}{>{\centering\arraybackslash}p{2cm}|>{\centering\arraybackslash}p{1.6cm}|>{\centering\arraybackslash}p{1.6cm}|>{\centering\arraybackslash}p{1.6cm}|>{\centering\arraybackslash}p{1.6cm}}
\hline  
$\tau$                    & 10 & 20 & 30 & 40  \\
\hline  
SST-2                   & 96.1  & 96.4 & 96.0  & 95.8  \\
MMLU                    & 85.8  & 86.0 & 85.6 & 85.3 \\
Avg. (↑)                    & 54.3  & 54.5 & 54.1 & 53.8 \\
\hline
\end{tabular}}
\vspace{-0.3cm}
\end{table}

\textbf{Effect of iteration frequency $\tau$.}\hspace{0.7em} \label{iteration frequency}
We explore the impact of different iteration frequencies, \ie, the number of local fine-tuning iterations before each round of pruning, on the final
model performance. 
As shown in Table \ref{table: iteration frequency}, the experimental results indicate that our default frequency ($\tau$ = 20) achieves the best performance.
Additionally, when pruning is too frequent ($\tau$ = 10), we observe that the model may not have enough iterations to recover through fine-tuning, resulting in inaccurate importance estimation. 
Furthermore, excessive fine-tuning between pruning iterations
($\tau$ = 40) causes the globally updated LoRA weights to drift toward the local data and thus deteriorates the model performance.



\section{Related Works}\label{sec:relatedwork}

\setlength{\parindent}{0pt} 
\textbf{Federated Fine-tuning of LLMs.}\hspace{0.7em}
LLMs have shown remarkable success across various domains, including CV, NLP, \etc
While on-device fine-tuning of LLMs can improve the performance on downstream tasks, privacy concerns limit the collection of distribution data from end devices. Federated Learning (FL) \cite{mcmahan2017communication} offers a promising solution to this challenge. 
For instance, FedNLP \cite{lin2021fednlp} provides a benchmarking framework to evaluate FL methods on NLP tasks. 
To address resource constraints on devices, most recent efforts
resort to Parameter-Efficient Fine-Tuning (PEFT) technique, among which 
LoRA \cite{hu2021lora} and Adapter \cite{houlsby2019parameter, karimi2021compacter, pfeiffer2020adapterfusion} are the most popular methods.
For example, FedAdapter \cite{cai2022fedadapter} utilizes adapters to accelerate model convergence in FL.   
FedPipe \cite{fang2024automated} leverages low-rank adapters \cite{hu2021lora}
to improve training speed and accuracy. 
FeDeRA \cite{yan2024federa} adopts singular value decomposition to further enhance the efficiency of LoRA-based fine-tuning. 

\textbf{Pruning of LLMs.}\hspace{0.7em}
LLM pruning can generally be categorized into structured and unstructured pruning. 
Unstructured pruning \cite{frantar2023sparsegpt, sun2023simple} prunes sparse parameters in LLMs,
which requires dedicated hardware support for efficiency improvements.
Meanwhile, structured pruning \cite{xia2022structured, ma2023llm, fang2023depgraph} removes groups of parameters (\eg, MHA heads, FFN neurons, and model dimensions) for ubiquitous inference efficiency gains in a hardware-agnostic way.
LLMPruner \cite{ma2023llm} combines first-order term and Hessian information for structured pruning.
Structured pruning of hidden dimensions, as shown by \cite{tao2023structured}, extends to embeddings and attention heads.
Sherared-LLaMA \cite{xia2023sheared} introduces a mask learning phase to identify prunable components.

\textbf{Combining Pruning and PEFT.}\hspace{0.7em}
Combining pruning and PEFT has the potential to improve both fine-tuning and inference efficiency.
CPET\cite{zhao2023cpet} applies different task-agnostic model compression methods to models fine-tuned with LoRA, but the performance degradation becomes notable specifically when structured pruning is employed.
Similarly, PST \cite{li2022parameter} and DSEE \cite{chen2021dsee} combine unstructured pruning with efficient fine-tuning, which hardly achieves acceleration on practical hardware.
LoRAPrune \cite{zhang2023loraprune} and APT \cite{zhao2024apt} integrate structured pruning of the LLMs with PEFT.
However, these methods are tailored for centralized cloud-based scenarios, which cannot be applied to end devices in FL with limited resources. 
For instance, APT requires calculating the product of activations and their gradients to estimate the weight importance, which incurs substantial computational and memory overhead.
Furthermore, arbitrarily applying these methods to FL settings must consider one crucial factor, \ie, device heterogeneity.



\section{Conclusion}\label{sec:concluesion}
In this paper, we design and implement \method, a novel FL framework that combines structured pruning and PEFT to enhance the fine-tuning and inference efficiency of on-device LLMs. 
Specifically, we iteratively prune and tune the parameters of LLMs,  achieving significant improvements in both fine-tuning and inference performance.
We then propose an MAB-based online learning algorithm to adaptively determine pruning ratios and LoRA ranks for different devices to conquer their heterogeneity.
Extensive experiments reveal that \method outperforms baselines in both convergence performance and speed, providing a speedup of 1.4-6.9$\times$ with the improvement of final model accuracy by 0.4\%-4.5\%.

\balance
\bibliographystyle{unsrt}
\bibliography{contents/refs}

\begin{thebibliography}{10}

\bibitem{cao2023comprehensive}
Yihan Cao, Siyu Li, Yixin Liu, Zhiling Yan, Yutong Dai, Philip~S Yu, and Lichao Sun.
\newblock A comprehensive survey of ai-generated content (aigc): A history of generative ai from gan to chatgpt.
\newblock {\em arXiv preprint arXiv:2303.04226}, 2023.

\bibitem{gunter2024apple}
Tom Gunter, Zirui Wang, Chong Wang, Ruoming Pang, Andy Narayanan, Aonan Zhang, Bowen Zhang, Chen Chen, Chung-Cheng Chiu, David Qiu, et~al.
\newblock Apple intelligence foundation language models.
\newblock {\em arXiv preprint arXiv:2407.21075}, 2024.

\bibitem{voigt2017eu}
Paul Voigt and Axel Von~dem Bussche.
\newblock The eu general data protection regulation (gdpr).
\newblock {\em A Practical Guide, 1st Ed., Cham: Springer International Publishing}, 10(3152676):10--5555, 2017.

\bibitem{mcmahan2017communication}
Brendan McMahan, Eider Moore, Daniel Ramage, Seth Hampson, and Blaise~Aguera y~Arcas.
\newblock Communication-efficient learning of deep networks from decentralized data.
\newblock In {\em Artificial intelligence and statistics}, pages 1273--1282. PMLR, 2017.

\bibitem{wang2018edge}
Shiqiang Wang, Tiffany Tuor, Theodoros Salonidis, Kin~K Leung, Christian Makaya, Ting He, and Kevin Chan.
\newblock When edge meets learning: Adaptive control for resource-constrained distributed machine learning.
\newblock In {\em IEEE INFOCOM 2018-IEEE conference on computer communications}, pages 63--71. IEEE, 2018.

\bibitem{xu2024fwdllm}
Mengwei Xu, Dongqi Cai, Yaozong Wu, Xiang Li, and Shangguang Wang.
\newblock Fwdllm: Efficient fedllm using forward gradient.
\newblock {\em arXiv. Available at: hjp://arxiv. org/abs/2308.13894 (Accessed: 11 March 2024)}, 2024.

\bibitem{houlsby2019parameter}
Neil Houlsby, Andrei Giurgiu, Stanislaw Jastrzebski, Bruna Morrone, Quentin De~Laroussilhe, Andrea Gesmundo, Mona Attariyan, and Sylvain Gelly.
\newblock Parameter-efficient transfer learning for nlp.
\newblock In {\em International conference on machine learning}, pages 2790--2799. PMLR, 2019.

\bibitem{hu2021lora}
Edward~J Hu, Yelong Shen, Phillip Wallis, Zeyuan Allen-Zhu, Yuanzhi Li, Shean Wang, Lu~Wang, and Weizhu Chen.
\newblock Lora: Low-rank adaptation of large language models.
\newblock {\em arXiv preprint arXiv:2106.09685}, 2021.

\bibitem{zhang2023fedpetuning}
Zhuo Zhang, Yuanhang Yang, Yong Dai, Qifan Wang, Yue Yu, Lizhen Qu, and Zenglin Xu.
\newblock Fedpetuning: When federated learning meets the parameter-efficient tuning methods of pre-trained language models.
\newblock In {\em Annual Meeting of the Association of Computational Linguistics 2023}, pages 9963--9977. Association for Computational Linguistics (ACL), 2023.

\bibitem{han2015deep}
Song Han, Huizi Mao, and William~J Dally.
\newblock Deep compression: Compressing deep neural networks with pruning, trained quantization and huffman coding.
\newblock {\em arXiv preprint arXiv:1510.00149}, 2015.

\bibitem{sanh2020movement}
Victor Sanh, Thomas Wolf, and Alexander Rush.
\newblock Movement pruning: Adaptive sparsity by fine-tuning.
\newblock {\em Advances in neural information processing systems}, 33:20378--20389, 2020.

\bibitem{ma2023llm}
Xinyin Ma, Gongfan Fang, and Xinchao Wang.
\newblock Llm-pruner: On the structural pruning of large language models.
\newblock {\em Advances in neural information processing systems}, 36:21702--21720, 2023.

\bibitem{kwon2022fast}
Woosuk Kwon, Sehoon Kim, Michael~W Mahoney, Joseph Hassoun, Kurt Keutzer, and Amir Gholami.
\newblock A fast post-training pruning framework for transformers.
\newblock {\em Advances in Neural Information Processing Systems}, 35:24101--24116, 2022.

\bibitem{xia2022structured}
Mengzhou Xia, Zexuan Zhong, and Danqi Chen.
\newblock Structured pruning learns compact and accurate models.
\newblock {\em arXiv preprint arXiv:2204.00408}, 2022.

\bibitem{zhao2024apt}
Bowen Zhao, Hannaneh Hajishirzi, and Qingqing Cao.
\newblock Apt: Adaptive pruning and tuning pretrained language models for efficient training and inference.
\newblock {\em arXiv preprint arXiv:2401.12200}, 2024.

\bibitem{zhao2023cpet}
Weilin Zhao, Yuxiang Huang, Xu~Han, Zhiyuan Liu, Zhengyan Zhang, and Maosong Sun.
\newblock Cpet: Effective parameter-efficient tuning for compressed large language models.
\newblock {\em arXiv preprint arXiv:2307.07705}, 2023.

\bibitem{molchanov2019importance}
Pavlo Molchanov, Arun Mallya, Stephen Tyree, Iuri Frosio, and Jan Kautz.
\newblock Importance estimation for neural network pruning.
\newblock In {\em Proceedings of the IEEE/CVF conference on computer vision and pattern recognition}, pages 11264--11272, 2019.

\bibitem{chen2022decentralized}
Suo Chen, Yang Xu, Hongli Xu, Zhida Jiang, and Chunming Qiao.
\newblock Decentralized federated learning with intermediate results in mobile edge computing.
\newblock {\em IEEE Transactions on Mobile Computing}, 23(1):341--358, 2022.

\bibitem{lai2021oort}
Fan Lai, Xiangfeng Zhu, Harsha~V Madhyastha, and Mosharaf Chowdhury.
\newblock Oort: Efficient federated learning via guided participant selection.
\newblock In {\em 15th $\{$USENIX$\}$ Symposium on Operating Systems Design and Implementation ($\{$OSDI$\}$ 21)}, pages 19--35, 2021.

\bibitem{fang2024automated}
Zihan Fang, Zheng Lin, Zhe Chen, Xianhao Chen, Yue Gao, and Yuguang Fang.
\newblock Automated federated pipeline for parameter-efficient fine-tuning of large language models.
\newblock {\em arXiv preprint arXiv:2404.06448}, 2024.

\bibitem{zhang2023loraprune}
Mingyang Zhang, Hao Chen, Chunhua Shen, Zhen Yang, Linlin Ou, Xinyi Yu, and Bohan Zhuang.
\newblock Loraprune: Pruning meets low-rank parameter-efficient fine-tuning.
\newblock {\em arXiv preprint arXiv:2305.18403}, 2023.

\bibitem{yang2024dual}
Yiyuan Yang, Guodong Long, Tao Shen, Jing Jiang, and Michael Blumenstein.
\newblock Dual-personalizing adapter for federated foundation models.
\newblock {\em arXiv preprint arXiv:2403.19211}, 2024.

\bibitem{yi2023fedlora}
Liping Yi, Han Yu, Gang Wang, and Xiaoguang Liu.
\newblock Fedlora: Model-heterogeneous personalized federated learning with lora tuning.
\newblock {\em arXiv preprint arXiv:2310.13283}, 2023.

\bibitem{wang2018glue}
Alex Wang.
\newblock Glue: A multi-task benchmark and analysis platform for natural language understanding.
\newblock {\em arXiv preprint arXiv:1804.07461}, 2018.

\bibitem{gao2020combinatorial}
Guoju Gao, Jie Wu, Mingjun Xiao, and Guoliang Chen.
\newblock Combinatorial multi-armed bandit based unknown worker recruitment in heterogeneous crowdsensing.
\newblock In {\em IEEE INFOCOM 2020-IEEE Conference on Computer Communications}, pages 179--188. IEEE, 2020.

\bibitem{fang2023depgraph}
Gongfan Fang, Xinyin Ma, Mingli Song, Michael~Bi Mi, and Xinchao Wang.
\newblock Depgraph: Towards any structural pruning.
\newblock In {\em Proceedings of the IEEE/CVF conference on computer vision and pattern recognition}, pages 16091--16101, 2023.

\bibitem{zhang2023adalora}
Qingru Zhang, Minshuo Chen, Alexander Bukharin, Nikos Karampatziakis, Pengcheng He, Yu~Cheng, Weizhu Chen, and Tuo Zhao.
\newblock Adalora: Adaptive budget allocation for parameter-efficient fine-tuning.
\newblock {\em arXiv preprint arXiv:2303.10512}, 2023.

\bibitem{cho2023heterogeneous}
Yae~Jee Cho, Luyang Liu, Zheng Xu, Aldi Fahrezi, Matt Barnes, and Gauri Joshi.
\newblock Heterogeneous lora for federated fine-tuning of on-device foundation models.
\newblock In {\em International Workshop on Federated Learning in the Age of Foundation Models in Conjunction with NeurIPS 2023}, 2023.

\bibitem{lai2022fedscale}
Fan Lai, Yinwei Dai, Sanjay Singapuram, Jiachen Liu, Xiangfeng Zhu, Harsha Madhyastha, and Mosharaf Chowdhury.
\newblock Fedscale: Benchmarking model and system performance of federated learning at scale.
\newblock In {\em International conference on machine learning}, pages 11814--11827. PMLR, 2022.

\bibitem{naik2016building}
Nitin Naik.
\newblock Building a virtual system of systems using docker swarm in multiple clouds.
\newblock In {\em 2016 IEEE International Symposium on Systems Engineering (ISSE)}, pages 1--3. IEEE, 2016.

\bibitem{paszke2019pytorch}
Adam Paszke, Sam Gross, Francisco Massa, Adam Lerer, James Bradbury, Gregory Chanan, Trevor Killeen, Zeming Lin, Natalia Gimelshein, Luca Antiga, et~al.
\newblock Pytorch: An imperative style, high-performance deep learning library.
\newblock {\em Advances in neural information processing systems}, 32, 2019.

\bibitem{gabriel2004open}
Edgar Gabriel, Graham~E Fagg, George Bosilca, Thara Angskun, Jack~J Dongarra, Jeffrey~M Squyres, Vishal Sahay, Prabhanjan Kambadur, Brian Barrett, Andrew Lumsdaine, et~al.
\newblock Open mpi: Goals, concept, and design of a next generation mpi implementation.
\newblock In {\em Recent Advances in Parallel Virtual Machine and Message Passing Interface: 11th European PVM/MPI Users’ Group Meeting Budapest, Hungary, September 19-22, 2004. Proceedings 11}, pages 97--104. Springer, 2004.

\bibitem{tirumala1999iperf}
Ajay Tirumala.
\newblock Iperf: The tcp/udp bandwidth measurement tool.
\newblock {\em http://dast. nlanr. net/Projects/Iperf/}, 1999.

\bibitem{liu2019roberta}
Yinhan Liu, Myle Ott, Naman Goyal, Jingfei Du, Mandar Joshi, Danqi Chen, Omer Levy, Mike Lewis, Luke Zettlemoyer, and Veselin Stoyanov.
\newblock Roberta: A robustly optimized bert pretraining approach.
\newblock {\em arXiv preprint arXiv:1907.11692}, 2019.

\bibitem{conover2023free}
Mike Conover, Matt Hayes, Ankit Mathur, Jianwei Xie, Jun Wan, Sam Shah, Ali Ghodsi, Patrick Wendell, Matei Zaharia, and Reynold Xin.
\newblock Free dolly: Introducing the world’s first truly open instruction-tuned llm.
\newblock {\em Company Blog of Databricks}, 2023.

\bibitem{liang2022holistic}
Percy Liang, Rishi Bommasani, Tony Lee, Dimitris Tsipras, Dilara Soylu, Michihiro Yasunaga, Yian Zhang, Deepak Narayanan, Yuhuai Wu, Ananya Kumar, et~al.
\newblock Holistic evaluation of language models.
\newblock {\em arXiv preprint arXiv:2211.09110}, 2022.

\bibitem{xiao2024large}
Jie Xiao, Qianyi Huang, Xu~Chen, and Chen Tian.
\newblock Large language model performance benchmarking on mobile platforms: A thorough evaluation.
\newblock {\em arXiv preprint arXiv:2410.03613}, 2024.

\bibitem{dhar2024empirical}
Nobel Dhar, Bobin Deng, Dan Lo, Xiaofeng Wu, Liang Zhao, and Kun Suo.
\newblock An empirical analysis and resource footprint study of deploying large language models on edge devices.
\newblock In {\em Proceedings of the 2024 ACM Southeast Conference}, pages 69--76, 2024.

\bibitem{hsu2019measuring}
Tzu-Ming~Harry Hsu, Hang Qi, and Matthew Brown.
\newblock Measuring the effects of non-identical data distribution for federated visual classification.
\newblock {\em arXiv preprint arXiv:1909.06335}, 2019.

\bibitem{yao2024ferrari}
Zhiwei Yao, Jianchun Liu, Hongli Xu, Lun Wang, Chen Qian, and Yunming Liao.
\newblock Ferrari: A personalized federated learning framework for heterogeneous edge clients.
\newblock {\em IEEE Transactions on Mobile Computing}, 2024.

\bibitem{lin2021fednlp}
Bill~Yuchen Lin, Chaoyang He, Zihang Zeng, Hulin Wang, Yufen Huang, Christophe Dupuy, Rahul Gupta, Mahdi Soltanolkotabi, Xiang Ren, and Salman Avestimehr.
\newblock Fednlp: Benchmarking federated learning methods for natural language processing tasks.
\newblock {\em arXiv preprint arXiv:2104.08815}, 2021.

\bibitem{karimi2021compacter}
Rabeeh Karimi~Mahabadi, James Henderson, and Sebastian Ruder.
\newblock Compacter: Efficient low-rank hypercomplex adapter layers.
\newblock {\em Advances in Neural Information Processing Systems}, 34:1022--1035, 2021.

\bibitem{pfeiffer2020adapterfusion}
Jonas Pfeiffer, Aishwarya Kamath, Andreas R{\"u}ckl{\'e}, Kyunghyun Cho, and Iryna Gurevych.
\newblock Adapterfusion: Non-destructive task composition for transfer learning.
\newblock {\em arXiv preprint arXiv:2005.00247}, 2020.

\bibitem{cai2022fedadapter}
Dongqi Cai, Yaozong Wu, Shangguang Wang, Felix~Xiaozhu Lin, and Mengwei Xu.
\newblock Fedadapter: Efficient federated learning for modern nlp.
\newblock {\em arXiv preprint arXiv:2205.10162}, 2022.

\bibitem{yan2024federa}
Yuxuan Yan, Qianqian Yang, Shunpu Tang, and Zhiguo Shi.
\newblock Federa: Efficient fine-tuning of language models in federated learning leveraging weight decomposition.
\newblock {\em arXiv preprint arXiv:2404.18848}, 2024.

\bibitem{frantar2023sparsegpt}
Elias Frantar and Dan Alistarh.
\newblock Sparsegpt: Massive language models can be accurately pruned in one-shot.
\newblock In {\em International Conference on Machine Learning}, pages 10323--10337. PMLR, 2023.

\bibitem{sun2023simple}
Mingjie Sun, Zhuang Liu, Anna Bair, and J~Zico Kolter.
\newblock A simple and effective pruning approach for large language models.
\newblock {\em arXiv preprint arXiv:2306.11695}, 2023.

\bibitem{tao2023structured}
Chaofan Tao, Lu~Hou, Haoli Bai, Jiansheng Wei, Xin Jiang, Qun Liu, Ping Luo, and Ngai Wong.
\newblock Structured pruning for efficient generative pre-trained language models.
\newblock In {\em Findings of the Association for Computational Linguistics: ACL 2023}, pages 10880--10895, 2023.

\bibitem{xia2023sheared}
Mengzhou Xia, Tianyu Gao, Zhiyuan Zeng, and Danqi Chen.
\newblock Sheared llama: Accelerating language model pre-training via structured pruning.
\newblock {\em arXiv preprint arXiv:2310.06694}, 2023.

\bibitem{li2022parameter}
Yuchao Li, Fuli Luo, Chuanqi Tan, Mengdi Wang, Songfang Huang, Shen Li, and Junjie Bai.
\newblock Parameter-efficient sparsity for large language models fine-tuning.
\newblock {\em arXiv preprint arXiv:2205.11005}, 2022.

\bibitem{chen2021dsee}
Xuxi Chen, Tianlong Chen, Weizhu Chen, Ahmed~Hassan Awadallah, Zhangyang Wang, and Yu~Cheng.
\newblock Dsee: Dually sparsity-embedded efficient tuning of pre-trained language models.
\newblock {\em arXiv preprint arXiv:2111.00160}, 2021.

\end{thebibliography}

\end{document}